\documentclass[journal]{IEEEtran}

\usepackage{graphicx}
\usepackage{amsmath}

\usepackage{dblfloatfix}
\usepackage{booktabs}
\usepackage{float}
\usepackage{multirow}
\ifCLASSINFOpdf
\else
\fi
\begin{document}
\title{Single-stream CNN with Learnable\\ 
       Architecture for Multi-source \\
       Remote Sensing Data}

\author{Yi Yang, 
        Daoye Zhu, 
        Tengteng Qu, 
		Qiangyu Wang, 
		Fuhu Ren, 
		Chengqi Cheng
\thanks{This work is supported by the National Key Research and 
Development Programs of China (2018YFB0505300). Corresponding 
author: Yi Yang. }
\thanks{Yi Yang and Fuhu Ren are with Center for Data 
		Science, Peking University, Beijing 100871, P.R.China
		(email: pkuyangyi@pku.edu.cn; renfh@pku.edu.cn)}
\thanks{Daoye Zhu is with Center for Data 
		Science, Peking University, Beijing 100871, P.R.China, 
		and Lab of Interdisciplinary Spatial Analysis, University 
		of Cambridge, Cambridge CB3 9EP, United Kingdom 
		(email: zhudaoye@pku.edu.cn)}
\thanks{Tengteng Qu, Qiangyu Wang and Chengqi Cheng are with 
		College of Engineering, Peking University, Beijing 
		100871, P.R.China. (email: tengteng.qu@pku.edu.cn; 
		wangqy522@pku.edu.cn; ccq@pku.edu.cn)
		}
}

\maketitle

\begin{abstract}
In this paper, we propose an efficient and generalizable 
framework based on deep convolutional neural network (CNN) 
for multi-source remote sensing data joint classification. 
While recent methods are mostly based on 
multi-stream architectures, we 
use group convolution to construct equivalent network 
architectures efficiently within a single-stream network. 
Based on a recent technique called dynamic grouping convolution, 
We further propose a network module named SepDGConv, 
to make group 
convolution hyperparameters, and thus the overall network 
architecture, learnable during network training. 
In the experiments, 
the proposed method is applied to ResNet and UNet, and the 
adjusted networks are verified on three very diverse 
benchmark data sets (i.e., Houston2018 data, Berlin data, 
and MUUFL data). Experimental results demonstrate 
the effectiveness of the proposed single-stream CNNs, and 
in particular SepG-ResNet18 improves
the state-of-the-art classification 
overall accuracy (OA) on HS-SAR Berlin data set from $62.23\%$ to 
$68.21\%$. In the experiments we have two interesting 
findings. First, using DGConv generally reduces test OA 
variance. Second, multi-stream is harmful to model 
performance if imposed to the first few layers, but becomes 
beneficial if applied to deeper layers. Altogether, the 
findings imply that 
multi-stream architecture, instead of being a strictly 
necessary component in deep learning models for multi-source 
remote sensing data, essentially plays the role of model 
regularizer. Our code is publicly available at 
https://github.com/yyyyangyi/Multi-source-RS-DGConv. 
We hope our work can inspire novel research in 
the future.
\end{abstract}


% \fi
\IEEEpeerreviewmaketitle

\section{Introduction}

\IEEEPARstart{R}{emote} sensing (RS) plays an important role 
in earth 
observation, and supports applications such as environmental 
monitoring \cite{li2020review}, precision agriculture 
\cite{sishodia2020applications}, etc. One fundamental yet 
challenging task in RS is land-use/land-cover (LULC) 
classification, which aims to assign one semantic category 
to each pixel in an RS image acuqired over some region of 
interest. 

Nowadays, diverse sensor technologies allow to measure 
different aspects of scenes and objects from the air, 
including sensors for
multispectral (MS) optical imaging, hyperspectral (HS) imaging, 
synthetic aperture radar (SAR), and light detection and 
ranging (LiDAR). Different sensors bring diverse and complementary 
information \cite{xu2019advanced}. 
For example, MS optical imagery contains spatial 
information such as object shape and spatial relationship. 
HS data provides detailed spectral information of LULC and 
ground objects. While HS imagery cannot be used to 
differentiate objects composed of the same material, such as 
roofs and roads both made of concrete, LiDAR data
can capture elevation distribution and thus can be used to 
distinguish roofs from roads. And SAR data can provide 
additional structure information about Earth’s surface. 
Availability of multi-source, multi-modal RS data makes it 
possible to integrate rich information to improve LULC 
classification performance. And considerable efforts have 
been invested into the research of multi-source RS data 
joint analysis for LULC since recent years. 

%------------------------------------------------

\subsection{Related work}

Conventionally, multi-source RS data analysis workflow contains 
two phases: one feature extraction phase, and one feautre 
fusion phase. In feature extraction phase, different feature 
extractors are applied to different data modalities, while 
in feature fusion phase, high level features obtained from 
the previous phase are fused by certain algorithms and fed 
to LULC classifiers. For example, both 
\cite{pedergnana2012classification} \cite{khodadadzadeh2015fusion} 
extract morphological attribute profiles from HS and LiDAR 
data, and use feature stacking as a fusion technique. 
In \cite{rasti2017hyperspectral}, 
morphological extinction profiles are extracted seperately 
from HS and LiDAR data, and the features are further fused 
using orthogonal TV component analysis (OTVCA). In 
\cite{gu2015novel}, the authors extract manually engineered 
features from MS and LiDAR data, and use multiple kernel 
learning as a fusion strategy to train a support vector 
machine classifier. In \cite{hu2019mima}, MAPPER 
\cite{singh2007topological} is used as a feature extractor 
for MS optical and SAR data, and the features are further 
fused with manifold alignment. Note that, in the literature, 
data fusion also refers to a data processing technique 
that integrates multi-source data into one data modality, 
while in our paper we use this term to express the same 
meaning as "joint classification / analysis" of multi-source 
RS data. 

Meanwhile, deep learning (DL) \cite{lecun2015deep}, as one 
of the most notable 
advances in computer vision (CV) recently, has attracted attention 
from both CV and RS communities. In particular, convolutional 
neural network (CNN) \cite{lecun2010convolutional}
is a DL based model that significantly 
outperforms traditional methods in image classification 
and segmentation. Most of the data in RS is 
also presented in the form of images, and CNN has shown 
remarkable success in analyzing MS \cite{marmanis2015deep}
\cite{maggiori2016convolutional} \cite{yuan2020review}, 
HS \cite{hu2015deep} \cite{li2016hyperspectral}
\cite{lee2017going}, LiDAR \cite{he2018lidar} 
\cite{pan2020land}, and SAR data \cite{zhao2016convolutional}
\cite{ma2018ship}. 

Besides being applied to individual data sources, CNNs are 
adopted as backbone models for multi-source RS 
data classification in many recent works. As an early attempt, 
\cite{chen2017deep} designs a two-branch CNN for joint 
analysis of HS-LiDAR data, one branch for each sensor, 
achieving promising classification accuracy. \cite{xu2017multisource} 
proposes another two-branch CNN for HS-LiDAR data, with a 
different design of HS feature extraction branch. In 
\cite{xu2018multi}, a three-branch CNN is proposed to fuse 
MS, HS and LiDAR data. \cite{hong2020more} further extend 
the scope of deep multi-branch networks by allowing either 
CNN or fully connected neural networks be one feature extraction 
branch. See Sec. II(A) for a formal definition of network branch. 

Based on the multi-branch architecture, 
some latest papers devote to further improve model performance 
by introducing various novel modules to the network. In \cite{hong2020x}, 
the authors use self-adversarial modules, interactive learning 
modules and label propagation modules to build a deep CNN 
for semi-supervised multi-modal learning. In \cite{zhang2021information}, 
Gram matrices are utilized to improve multisource 
complementary information preservation in a two-branch CNN 
for HS and LiDAR data fusion. In \cite{zhao2021fractional}, 
a two-branch CNN for joint classification of HS and LiDAR is 
proposed, where in the feature extraction branches Octave 
convolutional layers are used to reduce feature redundancy 
from low-frequency data components, and in the fusion 
sub-network fractional Gabor convolution is utilized to 
obtain multiscale and multidirectional spatial features. 

Using multi-stream architectures as above, this data fusion 
strategy is also known as the "late-fusion" scheme, since 
the features are kept sensor-specific until the last few 
layers, i.e., the fusion sub-network. Another strategy 
in contrast to late-fusion is "early-fusion". 
As the name suggests, early-fusion means either the feature 
extraction branches are very shallow, or there are certain 
information exchange between different branches, making the 
features no longer absolutely sensor-specific. 
In \cite{hang2020classification}, a coupled 
CNN is proposed to fuse HS and LiDAR data, where weight sharing 
technique is utilized in intermediate layers of the proposed 
two-branch CNN, making each feature extraction branch 
only contain very few separate layers. \cite{hazirbas2016fusenet} 
proposes a fully convolutional neural network named FuseNet 
with two 
branches for semantic labeling of indoor scenes on RGB-D data. 
To fuse features from RGB branch and depth branch, the authors 
propose a fusion block, which adds the output of each block 
from depth branch to RGB branch. In \cite{audebert2018beyond} 
a more detained comparison between FuseNet based early fusion 
and multi-branch late fusion is made, and the authors find 
out that one strategy does not consistently outperform the other 
on different data sets. 
In \cite{chen2019effective}, MS optical data 
and a digital surface model (DSM) band is jointly classified 
within a single-stream CNN with depth-wise convolution. 

%------------------------------------------------

\subsection{Challenges}

It can be summarized from the above literature that when 
designing a data fusion CNN, the following two principles 
are usually followed: (1) different branches are strictly 
separated, thus low level features are sensor-specific; 
(2) the number of branches is set equal or 
proportional to the number of data sources. 

Despite achieved success, it is far from fully understood 
why these empirical principles work. In particular, it may 
be helpful to further improve data fusion models, if we 
can gain some insight into the following two problems: 

\subsubsection{How to find optimal number of branches?}

Model performance and efficiency are both closely related to 
the number of branches, which is often treated as hyperparameters 
and defined by human 
experts. This can very likely lead the network to learn a 
sub-optimal solution, since experts cannot confidently know, 
and in fact there has been no agreement on, 
an optimal set of these hyperparameters. 
On the one hand, while it is possible to find an optimal number 
of branches by trial for small models, for typical 
modern CNNs which are very large 
($\sim 100$ layers \cite{khan2020survey}), manual tuning 
is no longer 
feasible. On the other hand, in a CNN, convolution layers in 
different depths learn features of different semantic 
meanings, and it can be very difficult to find an optimal 
depth that sensor-specific features are fused. 
It is therefore desirable that hyperparameters such as number 
of branches and branch depth can be found automatically. 

\subsubsection{Which works - specificity, or regularization?}

A multi-stream CNN has fewer parameters than its dense counterpart, 
for the latter there are more parameters to connect different branches. 
In the DL community, reducing model parameters is known as an 
effective regularization technique, which improves model 
performance. For multi-source data fusion, while it is generally 
assumed that sensor-specific features are beneficial, the 
effects of regularization has not been studied in isolation. 

\begin{figure*}[ht!]
	\begin{center}
			\includegraphics[width=1.7\columnwidth]{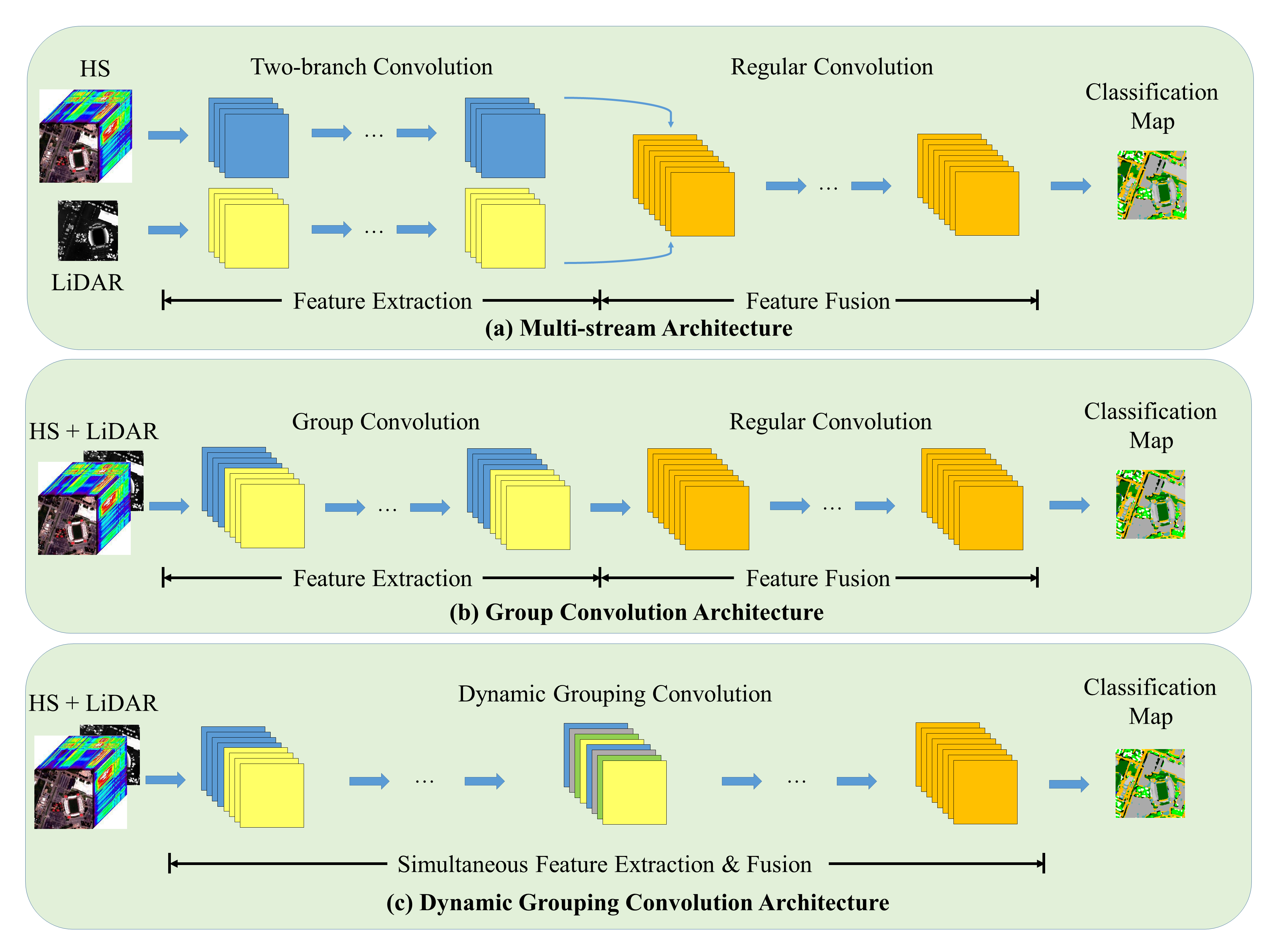}
		\caption{Illustration on the difference between 
		(a) multi-stream architecture, (b) group 
		convolution architecture, and (c) proposed 
		dynamic grouping convolution architecture.}
	\label{fig:overall}
	\end{center}
\end{figure*}

%------------------------------------------------

\subsection{Method overview}

To address the aforementioned challenges, we aim in this 
paper to develop a framework that allows CNN architectures 
to be learned from data within a single-stream 
network, for multi-source RS data fusion. 

We notice that any multi-stream architecture can be 
equivalently expressed by group convolution within a 
single-stream architecture (see Sec. II(A)). 
The following two parameters further control group convolution 
and need to be specified for each 
layer: number of total groups, and number of feature maps 
in each group. Recently proposed dynamic grouping convolution
\cite{zhang2019differentiable} (DGConv) enables these two 
parameters to be learned in an 
end-to-end manner via network training. 

Originally proposed 
for efficient architecture designing, DGConv itself does not ensure 
sensor-specific features and thus cannot be directly used to 
approximate and study multi-stream models for multi-source RS 
data fusion models. In our 
paper, we propose necessary modifications to DGConv, based on which 
we further design CNN blocks and single-stream 
architectures with simultaneous feature extraction-fusion 
for joint classification of multi-source RS data. 
Fig. 1(c) illustrates such a CNN model. 
More specifically, the contributions of this 
paper can be highlighted as follows.

(1) A modified DGConv module, which we name SepDGConv, 
is proposed to automatically 
learn group convolution structure within single-stream 
neural networks. 
SepDGConv is theoretically compatible with 
any CNN architecture. 

(2) Based on the proposed SepDGConv module, deep single-stream 
CNN models are proposed with reference to typical architectures 
in the CV area. The proposed CNNs show promising classification 
performance on 
various benchmark multi-source RS data sets. 

(3) Experimental results suggest that using densely connected 
network to jointly extract features from multiple data 
modalities actually improves the 
final classification performance, and using SepDGConv in 
deeper layers 
which contain more parameters helps improve classification 
accuracy. This finding is very 
interesting since it suggests that regularization contributes 
more to model performance improvement than sensor-specificity. 

(4) To our best knowledge, this is the first time that single-stream 
CNNs for multi-source RS data fusion is systematically studied 
and compared to state-of-the-art (SOTA) multi-branch models. 

The remainder of this paper is organized as follows. The
proposed DGConv layer and CNN architectures are introduced 
in Section II. The experimental results and analysis are 
presented in Section III. Finally,
Section IV makes the summary with some important 
conclusions and hints at potential future research trends.

%------------------------------------------------

\section{PRELIMINARIES}

In this section, firstly we show how we can use group convolution 
to construct a single-stream CNN that is equivalent to a 
multi-stream one. Secondly, we briefly introduce dynamic grouping 
convolution (DGConv) as well as G-Nets, a family of 
architectures using DGConv.

\subsection{Group conv as multi-stream conv}

A CNN branch/stream consists 
of a sequence of convolution/normalization/activation/pooling 
layers, and in a multi-stream architecture, the 
network branches usually play the role of feature extractor. 
Fig. 1(a) shows a typical two-branch CNN, 
with HS data fed to one branch (blue) and LiDAR to the other (yellow). 
The output features of multi-stream feature extractors are 
sensor-specific, since LiDAR data 
is never comes into the HS branch, and vice versa. These features 
are further fed into a fusion sub-network, which usually 
consists of one single branch. The fusion sub-network makes 
predictions based on the extracted features and outputs 
classification map. 

Group convolution, on the other hand, is originally proposed 
by the CV community for 
parallel computing \cite{krizhevsky2012imagenet}. This 
technique enables convolution in a layer to be computed 
in parallel groups. Group convolution is also studied in 
efficient network architecture designing 
\cite{xie2017aggregated}, since it reduces the number of 
parameters in convolution layers. Fig. 1(b) 
illustrates a single-stream CNN architecture using group 
convolution. Input HS and LiDAR data are stacked together. 
In the feature extraction sub-network, the 
feature maps are divided into two groups in parallel, 
blue and yellow. HS data goes into the yellow convolution 
group and LiDAR data goes into the blue group. 
Compared to Fig. 1(a), we can see that the two architectures 
are identical, suppose the architecture hyperparameters, 
i.e., number of layers, etc., are 
the same. 

Formally, Let 
$F \in \mathbf{R}^{N \times C^{in} \times H \times W}$ 
denote a feature map of a certain layer in a CNN, where $N, 
C^{in}, H, W$ represent minibatch size, number of input channels, 
height and width of the feature map, respectively. Let $\omega 
\in \mathbf{R}^{C^{out} \times C^{in} \times k \times k}$ be 
the convolution kernel in the same layer, where $C^{out}$ is 
the number of output channels and $k$ is kernel size. Then in 
a convolution operation $F$ and $\omega$ are multiplied to give 
an output feature map 
$O \in \mathbf{R}^{N \times C^{out} \times H \times W}$: 

\begin{equation}\label{equ:1}
	O(i, j) = \sum\limits_{m=0}^{k-1} \sum\limits_{n=0}^{k-1}
			  \omega(m, n) \, F(i+m, j+n)  
\end{equation}

where $i \in \{1, ..., H\}$, $j \in \{1, ..., W\}$, and 
$O(i, j) \in \mathbf{R}^{N \times C^{out}}$, 
$F(i+m, j+n) \in \mathbf{R}^{N \times C^{in}}$, 
$\omega(m, n) \in \mathbf{R}^{C^{out} \times C^{in}}$. 

In regular convolution, $\omega$ densely maps every input 
channel to every output channel, as illustrated in 
Fig. 2(a). In group convolution, such 
dense mapping is replaced by structured mapping such that, 
both feature maps and convolution kernels are divided into 
several groups on the channel dimension, and each kernel 
convolves only on feature maps 
in the same group as the kernel. Concretely, suppose we 
divide the convolution into $G$ groups,  

\begin{equation}\label{equ:2}
	O^{\gamma}(i, j) = \sum\limits_{m=0}^{k-1} 
		\sum\limits_{n=0}^{k-1} \omega^{\gamma}(m, n) 
		\, F^{\gamma}(i+m, j+n)
\end{equation}

\begin{tabbing} 
	where \hspace{0.1cm} \= $\gamma \in \{1, ..., G\}$\\
	\> $\cup$ represents concatenation along the channel axis\\
	\> $O(i, j) = O^1(i, j) \cup ... \cup O^{\gamma}(i, j) 
		\cup ... \cup O^G(i, j)$\\
	\> $\omega(i, j) = \omega^1(i, j) \cup ... \cup 
		\omega^{\gamma}(i, j) \cup ... \cup \omega^G(i, j)$\\
	\> $F(i, j) = F^1(i, j) \cup ... \cup F^{\gamma}(i, j) 
		\cup ... \cup F^G(i, j)$ 
\end{tabbing}

Fig. 2(c) illustrates a group convolution layer with 
$C^{in} = C^{out} = 8$ and $G = 2$. 

In the case of multi-source remote sensing data, if we 
assign one or multiple groups to each data modality, then 
we are able to embed multi-stream structure within a 
single-stream CNN. For example, suppose we have HS and LiDAR 
data, then we can set $G = 2$, and assign HS data to 
group 1 and LiDAR data to group 2. Using the illustration 
in Fig. 2(c), let's say HS data is on the blue colored group 
while LiDAR data is on the yellow colored group. Obviously
this is equivalent to a two-branch network with each branch 
having 4 channels. 

Furthermore, a common way to build deep sensor-specific 
branches is 
to set $G$ to a fixed vaule for all layers. Then in any 
layer, for any two different groups $\gamma_1$ 
and $\gamma_2$, $O^{\gamma_1}$ is computed only using 
$F^{\gamma_1}$, and $F^{\gamma_2}$ is never involved. 
This means that the output features are of the same 
sensor-specificity  
as the input, as the intuitive blue-to-blue, yellow-to-yellow 
illustration in Fig. 2(c). 

\begin{figure}[h!]
	\begin{center}
			\includegraphics[width=\columnwidth]{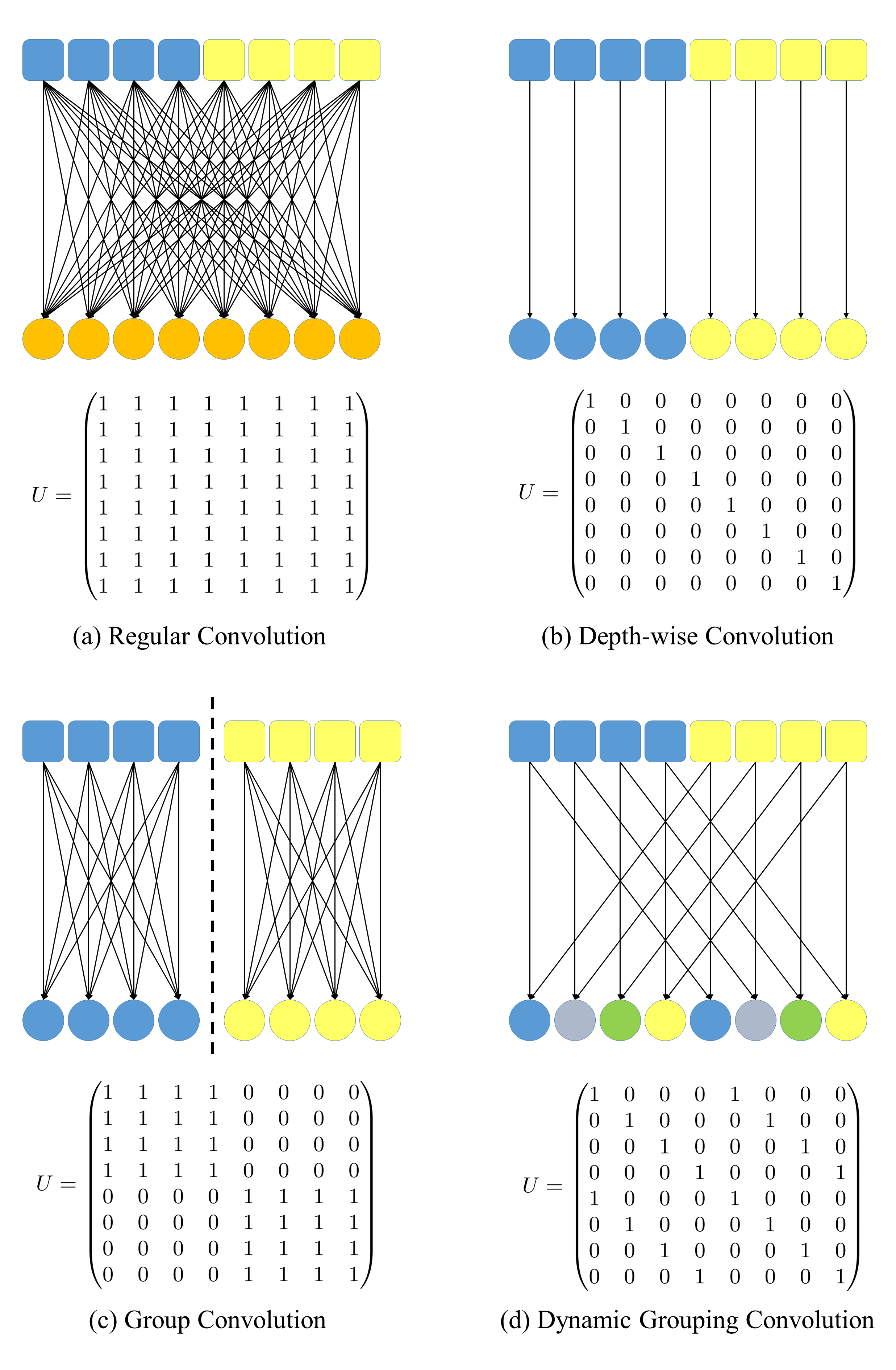}
		\caption{Illustration of different convolution 
		strategies (above) and their corresponding relationship 
		matrix U (below). One rectangle represents one input 
    feature map channel, and one circle represents one 
    output feature map channel. (a) regular convolution, (b) 
		depth-wise convolution, (c) group convolution, and 
		(d) dynamic grouping convolution.}
	\label{fig:gconv}
	\end{center}
\end{figure}

Existing models using group convolution may suffer from 
sub-optimal performance due to manually specifying hyperparameter 
$G$. To 
address this issue, DGConv allows both the total group number 
and channel connections to be learned from data, alongside with 
other CNN parameters. 

%------------------------------------------------

\subsection{Dynamic grouping convolution}

Here we briefly introduce dynamic grouping 
convolution (DGConv) \cite{zhang2019differentiable}. 
The key idea of DGConv is to model the input-output channel 
mapping by introducing a binary relationship matrix $U$, and 
then make $U$ learnable as part of the CNN parameters. 

\subsubsection{Definition} \quad

Formally, DGConv is defined as

\begin{equation}\label{equ:3}
	O(i, j) = \sum\limits_{m=0}^{k-1} \sum\limits_{n=0}^{k-1}
			  (U \odot \omega(m, n)) \, F(i+m, j+n)
\end{equation}

where $U \in \mathbf{R}^{C^{out} \times C^{in}}$, and $\odot$ 
denotes element-wise product. By this definition, $U_i$, 
the $i$th row of $U$, is a binary vector that indicates 
which input channels are involved in the computation of the 
$i$th output channel.The definition is reasonable, 
as many convolution operations can be regarded as special 
cases of DGConv. For instance, DGConv becomes regular convolution 
(Eq. (1)) if we let $U$ be a matrix of ones, as 
illustrated in Fig. 2(a). DGConv becomes 
depth-wise convolution if we let $U$ be an identity matrix, 
as illustrated in Fig. 2(b). 
DGConv can also represent group convolution (Eq. (2)), 
if we take $U$ 
to be a block-diagonal matrix of ones and zeros, 
as illustrated in Fig. 2(c). 

%------------------------------------------------

\subsubsection{Learning the relationship matrix U} \quad

While the definition in Eq. (3) is representative, such a 
definition results in the following two difficulties 
in estimating the value of $U$. First, the introduction 
of $U$ adds 
lots of additional parameters to the network, which makes 
the learning process more difficult. Second, $U$ takes 
binary values of 0 and 1, while it is widely known that 
optimization involving discrete values are generally very hard 
to solve. 

To address the first issue, $U$ is decomposed into a set of 
small matrices, and learnable parameters are designed to 
generate this set of small matrices. Consider a simple yet 
quite general case, where $U$ is a square matrix with $C^{in} 
= C^{out} = 2^K$, $K$ being an integer. Then a set of $2 
\times 2$ matrices $U_1, ..., U_i, ..., U_K$ can be defined, 
and $U$ can be reconstructed as 

\begin{equation}\label{equ:4}
	U = U_1 \otimes ... \otimes U_i \otimes ... \otimes U_K
\end{equation}

where $\otimes$ denotes Kronecker product, $i \in 
\{1, ..., K\}$. Each small matrix $U_i$ is further 
represented by a binary parameter $g_i \in \{0, 1\}$:

\begin{equation}\label{equ:5}
	U_i = g_i \mathbf{1} + (1-g_i) \mathit{I}
\end{equation}

where $\mathbf{1}$ denotes a $2 \times 2$ constant matrix of 
ones, $\mathit{I}$ denotes a $2 \times 2$ constant identity 
matrix. Thus each $2^K \times 2^K$ relationship matrix $U$ 
can be constructed from a vector $g \in \mathbf{R}^K$. The 
number of parameters to be learned is thereby 
reduced exponentially. 

To address the second issue, a learnable gate vector 
$\tilde{g}$, taking continuous values, is introduced to 
generate the binary vector $g$, as follows:

\begin{equation}\label{equ:6}
	g = \text{sign}(\tilde{g})
\end{equation}

where sign$(\cdot)$ represents the sign function

\begin{equation}\label{equ:7}
	\text{sign}(x)=\left\{
	\begin{aligned}
	& 0 & x < 0 \\
	& 1 & x \ge 0
	\end{aligned}
	\right.
\end{equation} 

Altogether, combining Eq. (4) - Eq. (7), the 
$2^K \times 2^K$ binary relationship matrix $U$ is 
constructed using a 
continuous vector $\tilde{g}$ of length $K$: 

\begin{equation}\label{equ:8}
	\begin{gathered}
	g = \text{sign}(\tilde{g}) \\
	U = g_1 \mathbf{1} + (1-g_1) \mathit{I} 
		\otimes ...  \otimes 
		g_K \mathbf{1} + (1-g_K) \mathit{I} 
	\end{gathered}
\end{equation}

Backpropagation through the non-differentiable sign($\cdot$) 
can be done by Straight-Through 
Estimator proposed for quantized neural networks \cite{yin2019understanding}
, and then  
automatic gradient computaton for the rest is supported by most 
modern deep learning programming frameworks.

As an example, a convolution layer with 
$C^{in} = C^{out} = 8$ as shown in Fig. (2) has $K = 3$, and 
the relationship matrix $U$ is of shape $8 \times 8$. 
$U$ for regular convolution in Fig. 2(a) can be expressed as 
$U = \mathbf{1} \otimes \mathbf{1} \otimes \mathbf{1}$, 
with $g = (1, 1, 1)$. Similarly, for depth-wise convolution 
in Fig. 2(b), $g = (0, 0, 0)$ and $U = \mathit{I} \otimes 
\mathit{I} \otimes \mathit{I}$. For group convolution 
illustrated as Fig. 2(c), $g = (1, 1, 0)$ and 
$U = \mathbf{1} \otimes \mathbf{1} \otimes \mathit{I}$. 
For DGConv shown in Fig. 2(d), $g = (0, 0, 1)$ and 
$U = \mathit{I} \otimes \mathit{I} \otimes \mathbf{1}$. 

%------------------------------------------------

\subsection{G-Nets}

Groupable-Networks (G-Nets) 
\cite{zhang2019differentiable} refer to architectures 
using DGConv. In particular, the authors of 
\cite{zhang2019differentiable} experiment with 
G-ResNet50, which is based on ResNet50 \cite{he2016deep}. 

ResNet50 uses Bottleneck as its building block. 
The Bottleneck block in order consists 
of one $1 \times 1$ convolution layer, one $3 \times 3$ 
convolution layer, and one more $1 \times 1$ convolution layer, 
as shown in Fig. 3(c). 
In its DGConv version, the middle $3 \times 3$ convolution 
is replaced with $3 \times 3$ DGConv. G-ResNet50 
consists of four Bottleneck blocks, the number of  
output channels for each block being $[256, 512, 1024, 2048]$, 
respectively. 

%------------------------------------------------

\section{METHOD}

While DGConv enables automatically learning of group 
convolution hyperparameters, the learning outcome does 
not lead to a network with sensor-specific branches, 
as we will see below, and 
thus cannot be directly used to approximate multi-branch 
CNNs for multi-source RS data fusion. To address this 
issue, based on DGConv we propose separable DGConv 
(SepDGConv) and separable G-Net (SepG-Net), which makes 
it possible that the learned architecture contains 
sensor-specific branches. 

\begin{figure}[h!]
	\begin{center}
			\includegraphics[width=\columnwidth]{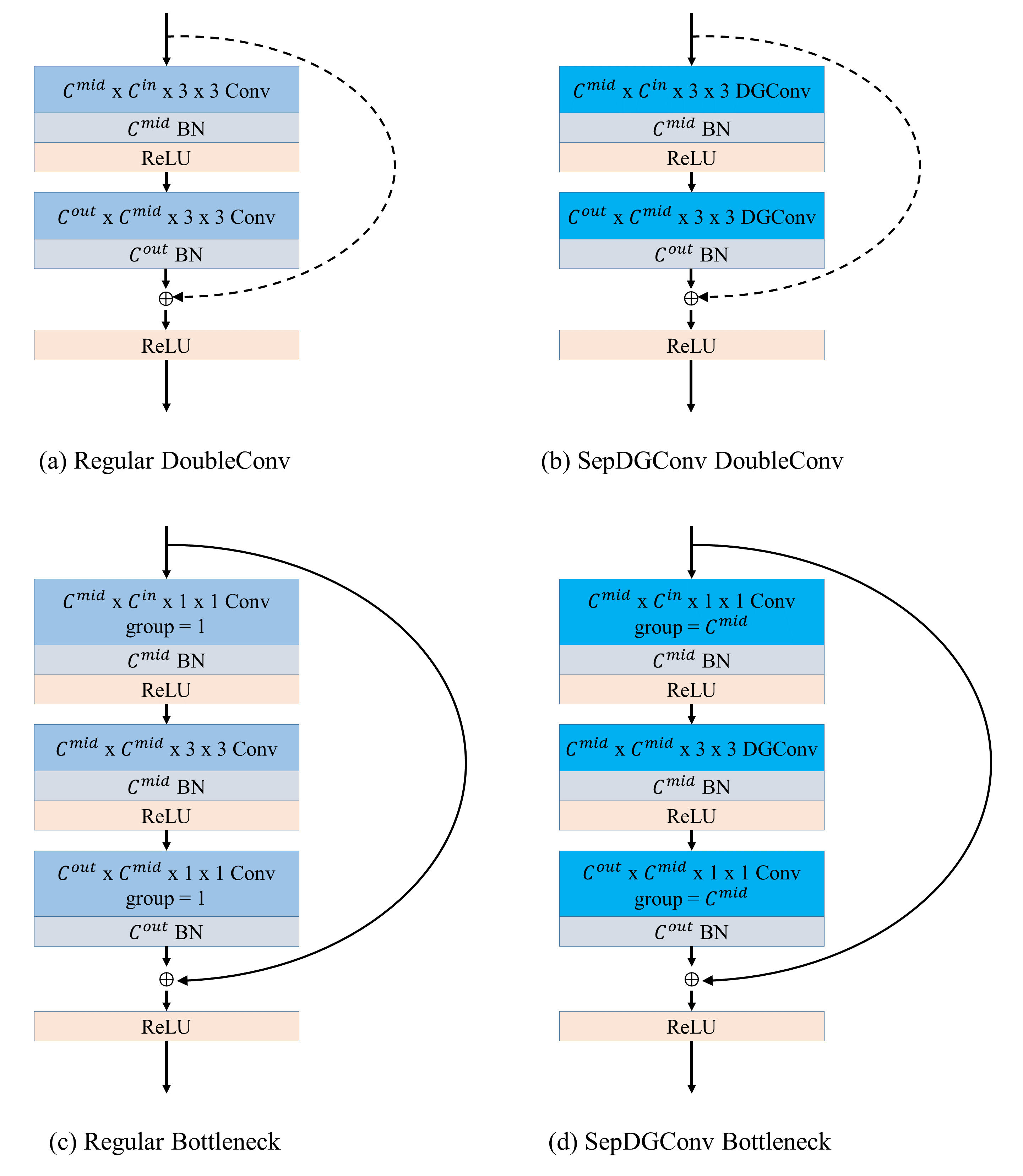}
		\caption{CNN blocks and their corresponding SepDGConv 
		variant. (a) regular DoubleConv, (b) SepDGConv DoubleConv 
		, (c) regular Bottleneck, and 
		(d) SepDGConv Bottleneck. The curve with arrow 
		stands for residual connection in ResNet blocks, and 
		$\oplus$ denotes summation. 
		In (a) and (b), the dashed curve means there is no 
		residual connection when DoubleConv is used in 
		UNet.}
	\label{fig:bloc}
	\end{center}
\end{figure}

\begin{figure*}[ht]
	\begin{center}
			\includegraphics[width=1.9\columnwidth]{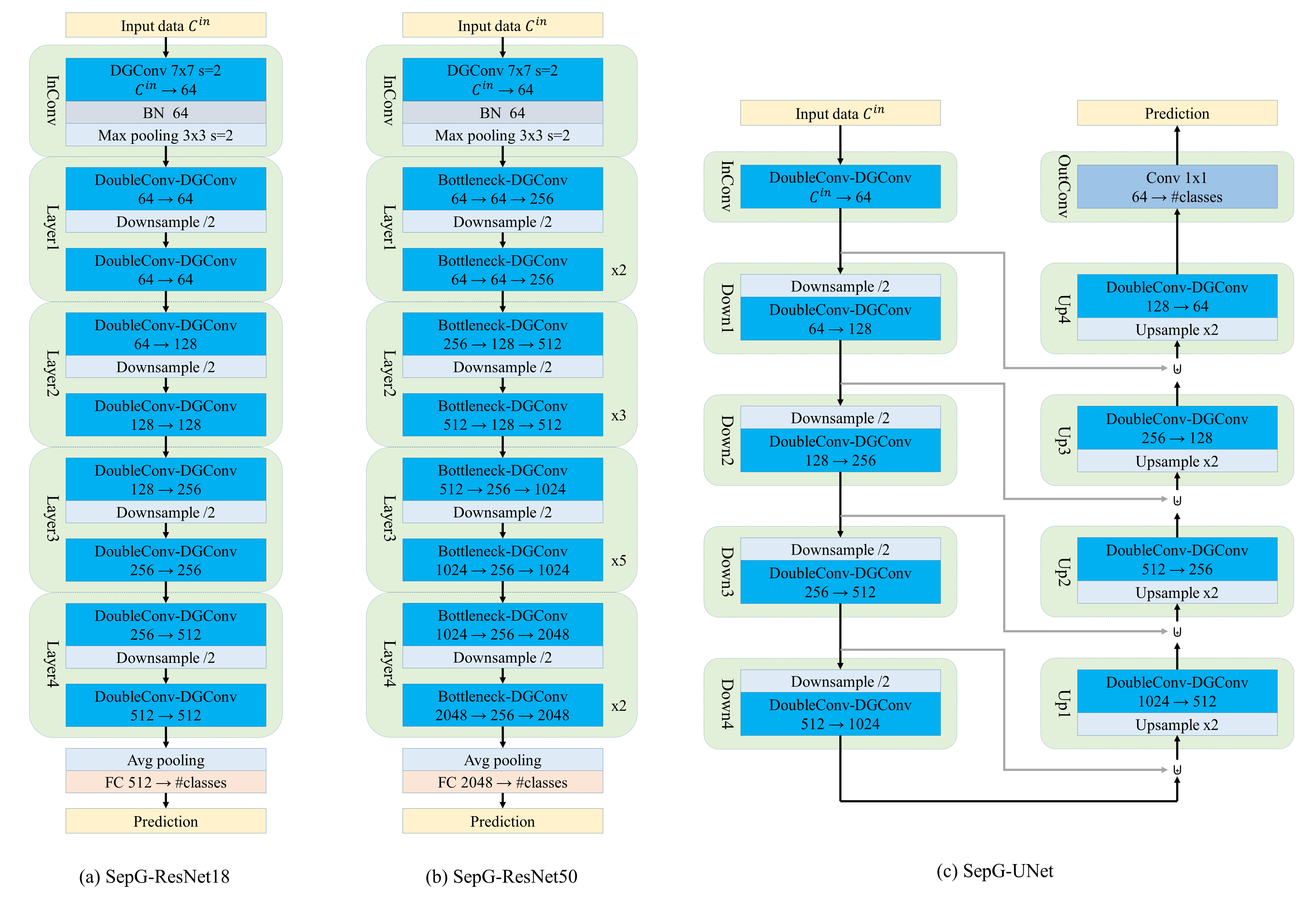}
		\caption{Neural network architectures using SepDGConv. 
		(a) SepG-ResNet18, (b) SepG-ResNet50, and (c) 
		SepG-UNet. In (b), the $\times x$ means that $x$ 
		consecutive blocks are omitted as one in the 
		illustration. 
		In (c), grey line represents skip connections 
		utilized in UNet, while $\uplus$ denotes 
		concatenation along the channel axis. 
		}
	\label{fig:arch}
	\end{center}
\end{figure*}

%------------------------------------------------

\subsection{Blocks with SepDGConv}

First, reconsider the Bottleneck block with DGConv. If 
sensor-specificity is to be preserved, then it is necessary 
that every convolution layer in a block uses group 
convolution, as in Eqn. (2). In both regular and DGConv 
Bottleneck blocks, the first and last convolution 
layer use $1 \times 1$ regular convolution. Therefore, 
in our SepDGConv Bottleneck, we use depth-wise convolution 
instead of regular convolution, i.e., we set the number of 
groups $G$ equal to the number of that layer's feature maps. 
Fig. 3(d) shows a SepDGConv Bottleneck block. 

Second, consider the DoubleConv block, which is used as 
building blocks by two very popular CNN models: ResNet18 
and UNet \cite{ronneberger2015u}. A DoubleConv block 
consists of two 
consecutive $3 \times 3$ convolution layers, as shown in 
Fig. 3(a). For the ResNet family, 
BasicBlock has a very similar structure to that of DoubleConv, 
except for the additional residual connection of BasicBlock. 
As long as there is no confusion, we also use 
DoubleConv to refer to BasicBlock in ResNets. To impose 
sensor-specificity, in our SepDGConv DoubleConv, we replace both 
convolution layers with DGConv layers, as shown in Fig. 3(b). 

%------------------------------------------------

\subsection{Details of SepDGConv}

Recall that in section II(B), for DGConv it is assumed on 
the relationship matrix $U \in \mathbf{R}^{C^{out} \times 
C^{in}}$ that $C^{out} = C^{in} = 2^K$. To apply DGConv 
to all layers in order to approximate deep sensor-specific 
branches, in our SepDGConv we handle the following two 
exceptions that violate the above assumption. 

\subsubsection{$C^{out} \neq C^{in}$} \quad

For architectures 
mentioned above, Bottleneck by our design satisfies this 
condition, however DoubleConv usually has $C^{out} \neq 
C^{in}$. $C^{out} > C^{in}$ is commonly seen in the feature 
extraction phase of a CNN, where the latter layer has 
more feature maps than its previous layer. $C^{out} < C^{in}$ 
is often used in upsampling layers of a segmentation model. 

Formally we define the following case \textit{expansion}: 
$C^{out} / C^{in} = r$, and the other case 
\textit{reduction}: $C^{in} / C^{out} = r$, with 
$r \ge 2$ being an integer. Our strategy is to expand or 
reduce the shape of the relationship matrix by doing 
matrix multiplication and Kronecker product with identity 
matrices and vectors of ones. 

In the \textit{expansion} case, we first construct a matrix 
$\tilde{U} \in \mathbf{R}^{C^{in} \times C^{in}}$ as described 
in subsection II(B). Recall that 
$U(i, j) = 1$ if the $j$th input channel is involved in the 
computation of the $i$th output channel, otherwise 
$U(i, j) = 0$. Hence we duplicate each 
row of $\tilde{U}$ $r$ times, and stack them together 
horizontally to get $U$: 

\begin{equation}\label{equ:9}
	U = (I \otimes \mathbf{1}_r) \tilde{U}
\end{equation}

where $I$ is a ${C^{in} \times C^{in}}$ identity matrix, 
and $\mathbf{1}_r$ is a column vector of ones, with length 
$r$. 

Similarly, in the \textit{reduction} case, we first construct 
$\tilde{U} \in \mathbf{R}^{C^{out} \times C^{out}}$. Then 
we duplicate each column of $\tilde{U}$ $r$ times, and 
stack them together vertically: 

\begin{equation}\label{equ:10}
	U = \tilde{U} (I \otimes \mathbf{1}^{\top}_r)
\end{equation}

where $I$ is a ${C^{out} \times C^{out}}$ identity matrix, 
and $\mathbf{1}^{\top}_r$ is a row vector of ones, with length 
$r$. 

With our proposed strategy we encourage feature maps in one 
input group to stay in the same output group. Also, note that 
both $(I \otimes \mathbf{1}_r)$ and 
$(I \otimes \mathbf{1}^{\top}_r)$ are fixed during network 
initialization phase and need no training. 

\subsubsection{$C^{in} \neq 2^K$} \quad

Finally we present how we handle $C^{in} \neq 2^K$, which 
is commonly seen in the input convolution 
block (InConv) of a CNN that computes convolution on the 
input data. 

For remote sensing data, the number of input data channels 
can range from $\sim 3$ (RGB data) to $\sim 10^2$ 
(HS data), while 
very commonly the number of input channels for the first 
block after InConv is set to $64$, which means for InConv 
$C^{out} = 64$. Hence $C^{in}$ for InConv can fall within 
any of the following 3 intervals: $(0, \frac{1}{2}C^{out})$, 
$(\frac{1}{2}C^{out}, 2C^{out})$, $(2C^{out}, +\infty)$. 

If $0 < C^{in} < \frac{1}{2}C^{out}$, then we hope to use 
Eqn(9) to construct $U$. We make the following modification 
to make sure both $K$ and $r$ are integers:

\begin{equation}\label{equ:11}
	\begin{gathered}
	K = \lceil \text{log}_2(C^{in}) \rceil \\
	r = \lceil C^{out} / C^{in} \rceil
	\end{gathered}
\end{equation}

where $\lceil x \rceil$ denotes the round up function. 

Similarly, we use Eqn(10) for $2C^{out} < C^{in} < +\infty$, 
with 

\begin{equation}\label{equ:12}
	\begin{gathered}
	K = \lceil \text{log}_2(C^{out}) \rceil \\
	r = \lceil C^{in} / C^{out} \rceil
	\end{gathered}
\end{equation}

For the last case $\frac{1}{2}C^{out} < C^{in} < 2C^{out}$, 
we determine $K$ and round it up to an integer: 

\begin{equation}\label{equ:13}
	K = \lceil \text{log}_2(\text{max}(C^{in}, C^{out})) \rceil
\end{equation}

where $\text{max}(a, b)$ takes the maximum of $a$ and $b$. 

Finally, the size of $U$ constructed by our design 
is always larger than $\mathbf{R}^{C^{out} \times C^{in}}$. 
Hence we only use the first $C^{out}$ rows and $C^{in}$ 
columns in our computation, ignoring the remaining entries. 

%------------------------------------------------

\subsection{SepG-Nets}

Using SepDGConv DoubleConv, we can build SepG-ResNet18, as shown 
in Fig. 4(a), and SepG-UNet, as shown in Fig. 4(c). 
In SepG-ResNet18, we follow the convention and build 
the network with four layers, each having 2 DoubleConv blocks. 
The number of  
output channels for each layer being $[64, 128, 256, 512]$, 
respectively. In SepG-UNet, we have 8 DoubleConv 
blocks, with output channel numbers $[128, 256, 512, 1024, 
512, 256, 128, 64]$. 

SepG-ResNet50 consists of four layers, in order composed of 
$[2, 3, 5, 2]$ SepDGConv Bottleneck blocks. 
The number of output channels for each block is the same as 
G-ResNet50, being $[256, 512, 1024, 2048]$ respectively. The 
architecture of SepG-ResNet50 is illustrated in Fig. 4(b). 

%------------------------------------------------

\section{EXPERIMENTS AND DISCUSSION}\label{sec:EXPERIMENT}

We experiment with SepG-ResNet18, SepG-ResNet50 and 
SepG-UNet on three diverse data sets: Houston2018 data set, 
Berlin data set and MUUFL data set, and compare with baseline 
models, as well as SOTA models on the 
data sets. In this section, firstly we describe the data 
sets. Secondly we present our experimental results. Thridly,
we analyze the role of SepDGConv in the entire model by ablation 
analysis and we report changes in classification performance. 
Fourthly, we compare different convolution strategies, 
in particular group convolution and SepDGConv, to isolate 
the effect of sensor-specificity. 
Finally, we discuss the results and findings of our 
experiments. 

\subsection{Data sets}

\subsubsection{Houston2018 data set} \quad

Houston2018 is an HS-LiDAR-RGB data set. 
Acuqired by the National Center for Airborne Laser Mapping 
at the University of Houston, the Houston2018 data set covers 
the University of Houston campus and its surrounding
urban areas. The data set consists of MS-LiDAR, 
HS, and MS-optical remote sensing data, each containing 7, 48, 
and 3 channels, respectively. The HS data covers a 
380–1050 nm spectral range, while laser wavelengths of 
3 LiDAR sensors are 1550, 1064 and 532 nm. The MS-LiDAR data 
also contains DEM and DSM derived from point clouds. 
This data set was originally 
provided in 2018 GRSS Data Fusion Contest. This 
paper \cite{xu2019advanced} reports the outcome of the 
Contest, and also 
contains more detailed description of the Houston2018 data 
set. We resample the imagery at a 0.5-m GSD, so that the 
size of each image channel is $2404 \times 8344$ pixels. The 
ground truth of data set contains 20 classes. The number 
of samples in each class are shown in Table I. 

\begin{table}[!h]
	\centering
	\caption{LULC classes in Houston2018 data set}
	\resizebox{\columnwidth}{!}
	{
	  \begin{tabular}{cc|c|c}
		\toprule
		\multicolumn{1}{c}{\# } & Class & \# Training samples & \# Test samples \\
		\midrule
		\multicolumn{1}{c}{1} & Healthy Grass & 39196 & 20000 \\
		\multicolumn{1}{c}{2} & Stressed Grass & 130008 & 20000 \\
		\multicolumn{1}{c}{3} & Artificial Turf & 2736  & 20000 \\
		\multicolumn{1}{c}{4} & Evergreen Trees & 54322 & 20000 \\
		\multicolumn{1}{c}{5} & Deciduous Trees & 20172 & 20000 \\
		\multicolumn{1}{c}{6} & Bare Earth & 18064 & 20000 \\
		\multicolumn{1}{c}{7} & Water & 1064  & 1628 \\
		\multicolumn{1}{c}{9} & Non-residential Buildings & 894769 & 20000 \\
		\multicolumn{1}{c}{10} & Roads & 183283 & 20000 \\
		\multicolumn{1}{c}{11} & Sidewalks & 136035 & 20000 \\
		\multicolumn{1}{c}{12} & Crosswalks & 6059  & 5345 \\
		\multicolumn{1}{c}{13} & Major Thoroughfares & 185438 & 20000 \\
		\multicolumn{1}{c}{14} & Highways & 39348 & 20000 \\
		\multicolumn{1}{c}{15} & Railways & 27748 & 11232 \\
		\multicolumn{1}{c}{16} & Paved Parking Lots & 45932 & 20000 \\
		\multicolumn{1}{c}{17} & Unpaved Parking Lots & 587   & 3524 \\
		\multicolumn{1}{c}{18} & Cars  & 26289 & 20000 \\
		\multicolumn{1}{c}{19} & Trains & 21479 & 20000 \\
		\multicolumn{1}{c}{20} & Stadium Seats & 27296 & 20000 \\
		\midrule
			  & Total & 1859825 & 321729 \\
		\bottomrule	
	  \end{tabular}%
	}
	\label{tab:ds_houston}%
  \end{table}%  

\subsubsection{Berlin data set} \quad

This is an HS-SAR data set. 
The Berlin data set covers the Berlin urban and its rural 
neighboring area. The data set consists of HS and MS-SAR  
imagery, containing 244 and 4 channels respectively. 
The HS data was originally provided by \cite{okujeni2016berlin}, 
with wavelength range of 400 nm to 2500 nm, while more 
recently the authors of \cite{hong2021multimodal} 
acquired MS-SAR data of the same 
area and processed both HS and MS-SAR data into analysis-
ready form, which is used in our experiments. The processed 
data has a 13.89m GSD and consists of $1723 \times 476$ 
pixels. The Berlin data set contains 15 classes of ground 
truth labels. The number 
of samples in each class are shown in Table II.

\begin{table}[htbp]
	\centering
	\caption{LULC classes in Berlin data set}
	\resizebox{\columnwidth}{!}
	{
	  \begin{tabular}{rc|c|c}
	  \toprule
	  \multicolumn{1}{c}{\# } & Class & \# Training samples & \# Test samples \\
	  \midrule
	  \multicolumn{1}{c}{1} & Forest & 443   & 54511 \\
	  \multicolumn{1}{c}{2} & Residential Area & 423   & 268219 \\
	  \multicolumn{1}{c}{3} & Industrial Area & 499   & 19067 \\
	  \multicolumn{1}{c}{4} & Low Plants & 376   & 58906 \\
	  \multicolumn{1}{c}{5} & Soil  & 331   & 17095 \\
	  \multicolumn{1}{c}{6} & Allotment & 280   & 13025 \\
	  \multicolumn{1}{c}{7} & Commercial Area & 298   & 24526 \\
	  \multicolumn{1}{c}{8} & Water & 170   & 6502 \\
	  \midrule
			& total & 2820  & 461851 \\
	  \bottomrule
	  \end{tabular}%
	}
	\label{tab:ds_berlin}%
  \end{table}%

\subsubsection{MUUFL Data Set} \quad

The MUUFL Gulfport data set \cite{du2017technical}
is an HS-LiDAR data set. It 
was collected over the University 
of Southern Mississippi Gulf Park Campus. The data set contains 
coregistered HS and MS-LiDAR data, with 64 and 2 bands 
respectively. The wavelength of HS data spectral bands ranges 
from 375 to 1050 nm. The data set contains $325 \times 220$ 
pixels, with spatial resolution of 0.54 m across track
and 1.0 m along track. In the ground truth labels there are 
11 classes. The number 
of samples in each class are shown in Table III.

\begin{table}[htbp]
	\centering
	\caption{LULC classes in MUUFL data set}
	\resizebox{\columnwidth}{!}
	{
	  \begin{tabular}{rc|c|c}
	  \toprule
	  \multicolumn{1}{c}{\# } & Class & \# Training samples & \# Test samples \\
	  \midrule
	  \multicolumn{1}{c}{1} & Trees & 100   & 23246 \\
	  \multicolumn{1}{c}{2} & Mostly Grass & 100   & 4270 \\
	  \multicolumn{1}{c}{3} & Mixed Ground Surface & 100   & 6882 \\
	  \multicolumn{1}{c}{4} & Dirt and Sand & 100   & 1826 \\
	  \multicolumn{1}{c}{5} & Road  & 100   & 6687 \\
	  \multicolumn{1}{c}{6} & Water & 100   & 466 \\
	  \multicolumn{1}{c}{7} & Building Shadow & 100   & 2233 \\
	  \multicolumn{1}{c}{8} & Building & 100   & 6240 \\
	  \multicolumn{1}{c}{9} & Sidewalk & 100   & 1385 \\
	  \multicolumn{1}{c}{10} & Yellow Curb & 100   & 183 \\
	  \multicolumn{1}{c}{11} & Cloth Panels & 100   & 269 \\
	  \midrule
			& Total & 1100  & 53687 \\
	  \bottomrule
	  \end{tabular}%
	}
	\label{tab:ds_muufl}%
  \end{table}%  

%------------------------------------------------

\subsection{Classification and results}

\begin{figure*}[htbp]
	\begin{center}
			\includegraphics[width=2\columnwidth]{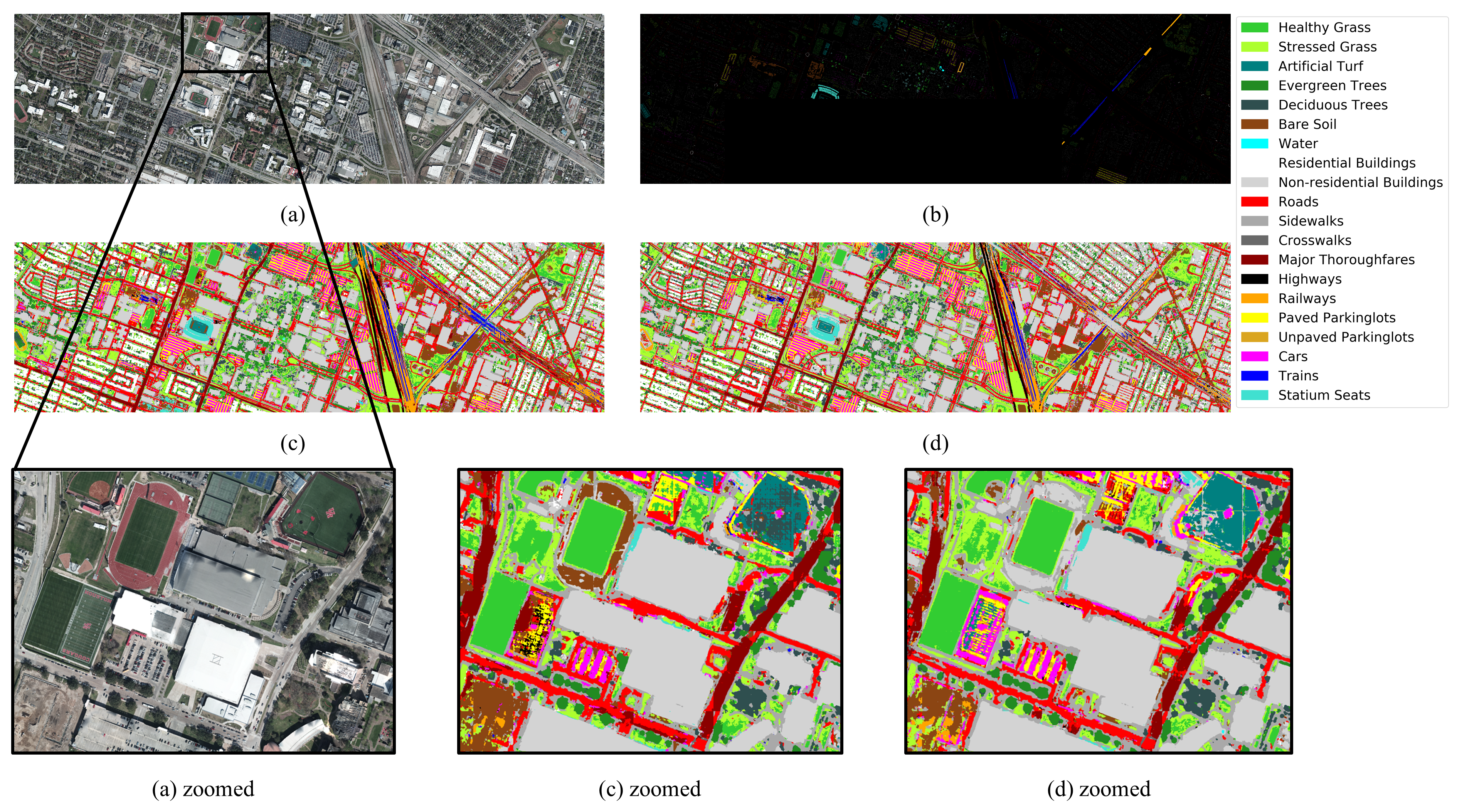}
		\caption{Results on Houston2018 data set. 
		(a) MS optical image, (b) test ground truth labels, 
		(c) prediction map by ResNet18, and (d) prediction 
		map by SepG-ResNet18. Zoomed in image from (a), (c) 
		and (d) are shown in the bottom row. 
		}
	\label{fig:pred_houston}
	\end{center}
\end{figure*}

Since the data modality of the above 3 data sets are quite 
different from each other, we use different classification 
models and compare with different methods on each data set. 
We will present data preparation, experimental settings and 
classification results seperately for each data set. Here 
we state some global configurations in our experiments. 

\textbf{Environment and reproducibility} We run our models 
on NVIDIA GTX 1080 Ti GPUs throughout our experiments. 
Software we use include: CUDA 10.2, cuDNN 7.6.5, python 
3.8.2, pytorch 1.9.0, torchvision 0.10.0, scipy 1.6.2, 
and numpy 1.20.2. In all our experiments we run 5 replicas 
for each model using random seeds 42-46. The 
reproducibility of our experimental results depends on 
random seeds, software, and hardware. Note that using 
deterministic algorithms to a certain extent reduces 
classification accuracy of our models. 

\textbf{Preprocessing} For all the data utilized in our 
experiments, we use channel-wise normalize to rescale each 
channel into the range $[0, 1]$: 

\begin{equation}\label{equ:14}
	channel[i, j] = \frac{channel[i, j] - \text{min}(channel)}
				   {\text{max}(channel) - \text{min}(channel)}
\end{equation}

where $channel$ is a 2D array representing one image channel, 
$\text{max}(channel)$ and $\text{min}(channel)$ takes the maximum and 
minimum value from $channel$, respectively. 

For more details of data preparation we follow previous work 
unless otherwise specified. For Houston2018 data set, we 
follow \cite{xu2018multi}. For Berlin data set, we follow 
\cite{hong2021multimodal}. For MUUFL data set, we follow 
\cite{zhao2021fractional}. 

\textbf{Initialization} If not specified, we draw initial 
values from the uniform distribution, which is also the 
default initialization method in Pytorch. 

\textbf{Optimization} Since there are varying degrees of 
data imbalance in all the three data sets, we use the weighted 
cross entropy loss to train CNN models, with weight for 
each class set to

\begin{equation}\label{equ:15}
	w_c = 1 - \frac{\#samples_c}
				   {\#total\_samples}
\end{equation}

where $w_c$ denotes class weight for class $c$, $\#samples_c$ 
represents number of training samples of class $c$, and 
$\#total\_samples$ represents the total number of 
samples in the training set. 

\textbf{Metrics} To 
evaluate the classification results, for each classifier 
we will report F1-score (F1) for each class, and three 
widely-used criteria in the literature to evaluate overall 
performance, i.e., average accuracy (AA), overall 
accuracy (OA) and Kappa coefficient ($\kappa$). 

%------------------------------------------------
\vspace{6pt}

\subsubsection{Experiments on Houston2018 data set} \quad

\begin{figure*}[htbp]
	\begin{center}
			\includegraphics[width=2\columnwidth]{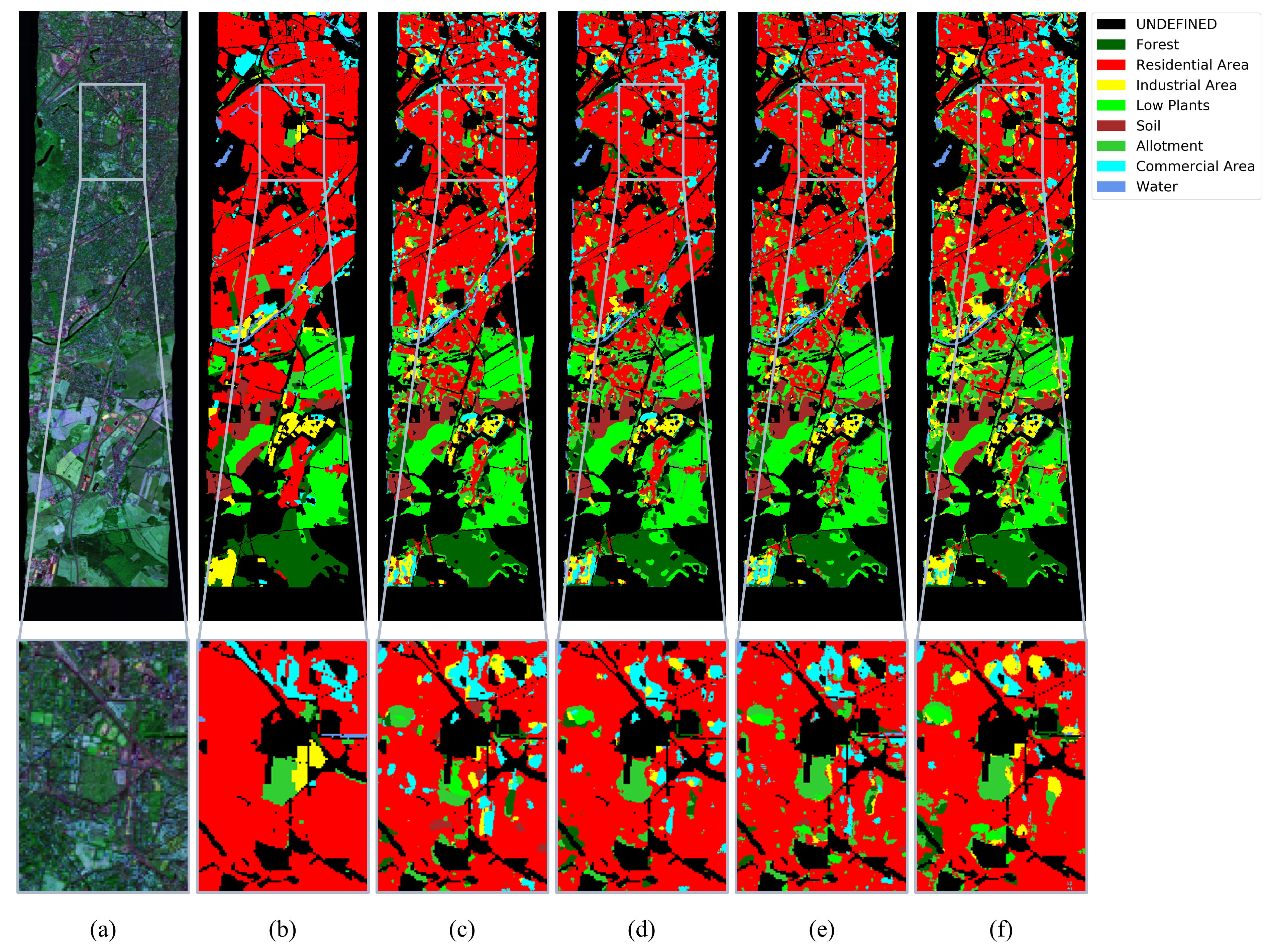}
		\caption{Results on Berlin data set. 
		(a) HS false color image, (b) test ground truth labels, 
		and prediction maps by 
		(c) ResNet18, (d) SepG-ResNet18, (e)  
		ResNet50, and (f) SepG-ResNet50. 
		}
	\label{fig:pred_berlin}
	\end{center}
\end{figure*}

We follow \cite{xu2018multi} and treat the LULC 
classification on Houston2018 data set as a semantic 
segmentation problem, and experiment with UNet, a 
very commonly studied semantic segmentation model. 
We run basic UNet on the data set as a baseline, 
and examine the performance of SepG-UNet, 
where SepDGConv layers replace regular convolution 
layers in the baseline model. We also compare with the 
first and second place methods presented in 2018 Data 
Fusion Contest, both are multi-stream models: 
Fusion-FCN \cite{xu2018multi} and DCNN 
\cite{cerra2018combining}. 

\textbf{Implementation details} We use Image tiles of shape 
$58 \times 128 \times 128$. In the training phase we use 
a spatial stride of $64 \times 64$ pixels to extract training 
samples from the $58 \times 1202 \times 4768$ data, while 
in the test phase the stride is $128 \times 128$. 

We use the Adam optimizer 
\cite{kingma2014adam} to train both UNet and SepG-UNet, 
with optimizer hyperparameters $\beta_1$ and $\beta_2$ set 
to default values. We set the initial learning rate to 0.001, 
and train the networks for 300 epochs. We use 3 GPUs in 
parallel to train the networks, with batch size set to 
12. To ensure reproducibility we use transpose convolution 
in upsampling modules in UNet and SepG-UNet. While the 
reference data for Houston2018 data set is densely labeled, 
there are still some areas on the image where there are no 
annotations. We add a mask to the loss function so that 
these undefined pixels are not accounted for training loss. 

\textbf{Results} The quantitative classification results of 
SepG-UNet, UNet, Fusion-FCN and DCNN are shown in Table
IV, while Fig. 5(c) and 5(d) shows the classification map of 
UNet and SepG-UNet, alongside with the ground truth 
labels. In Table IV, the classification 
results of DCNN and Fusion-FCN are directly cited from 
\cite{xu2019advanced}. Note that, in \cite{xu2019advanced} 
ensembling is used and the authors actually report the 
performance of an ensemble of Fusion-FCN. Our 
experimental results show that basic UNet, which 
does not has a multi-stream architecture, can 
yield an overall accuracy of $63.66\%$, 
outperforming $63.28\%$ obtained by ensembled Fusion-FCN, 
and largely outperforming DCNN. And with the proposed 
SepDGConv, the performance of SepG-UNet is further improved 
to $63.66\%$. From Fig. 5, it can be 
seen that both UNet and SepG-UNet can output meaningful 
prediction maps. According to Table IV, using SepDGConv reduces 
OA variance, which is reflected in Fig. 5(c) and Fig. 5(d)
that SepG-UNet gives a generally less noisy 
classification map than basic UNet.

In Fig. 8(a), we plot the learned number of groups and 
sparsity of relationship matrix $U$ for each SepDGConv layer 
in SepG-UNet. Here the sparsity of $U$ is defined as the 
ratio between number of 0s and number of total entries 
in $U$. The \textit{Sparsity} plot shows that SepDGConv generates 
an architecture with dense-sparse connection alternately 
appearing. 
While in InConv block the learned $U$s are mostly 
sparse, the sparsities quickly drop to 0 in Down1 block 
in all 5 replicas, which suggests that sensor-specific 
branches learned in SepG-UNet are very shallow, 
and feature fusion probably begins in a very early stage. 
In the \textit{\#Groups} plot, we can see that SepDGConv 
learns more groups in middle layers where there are more 
feature maps and more parameters. This is consistent with 
the behaviour of DGConv found in the paper 
\cite{zhang2019differentiable}. 

%------------------------------------------------
\vspace{6pt}

\subsubsection{Experiments on Berlin data set} \quad

\begin{figure*}[htbp]
	\begin{center}
			\includegraphics[width=2\columnwidth]{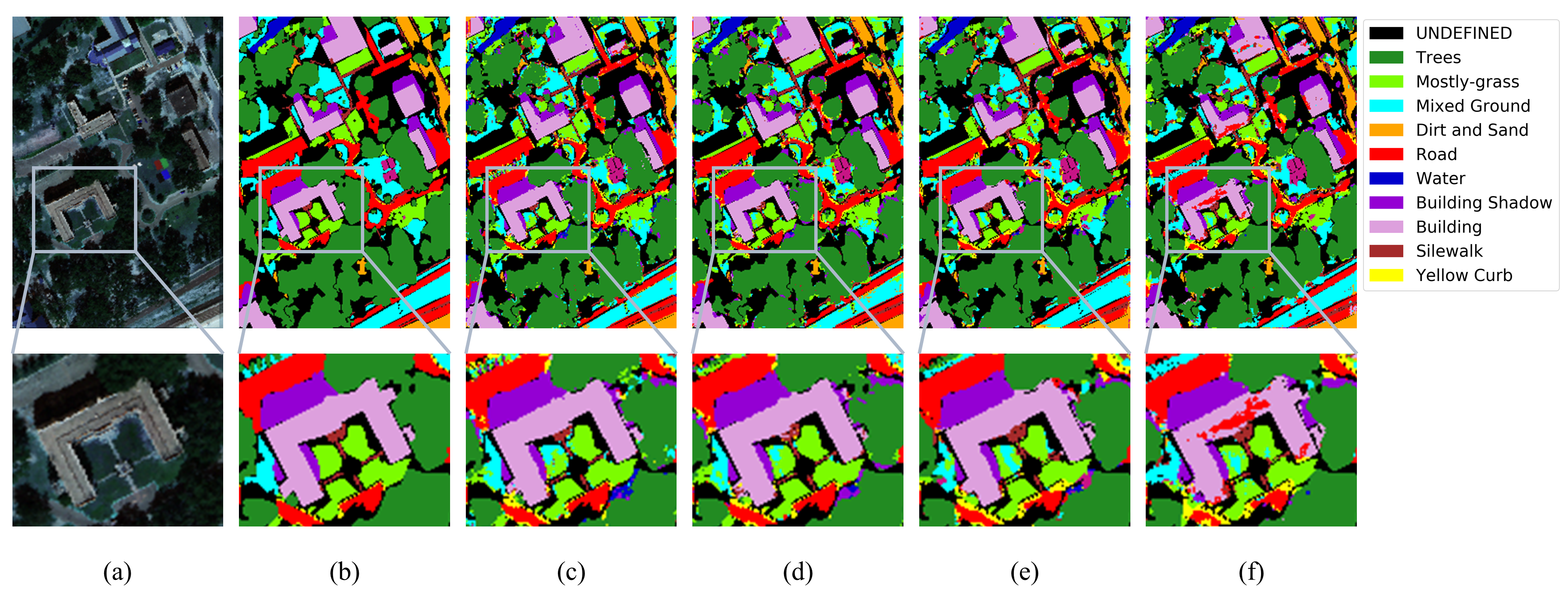}
		\caption{Results on MUUFL data set. 
		(a) HS false color image, (b) test ground truth labels, 
		and prediction maps by 
		(c) ResNet18, (d) SepG-ResNet18, (e) 
		ResNet50, (f) SepG-ResNet50. 
		}
	\label{fig:pred_muufl}
	\end{center}
\end{figure*}

In Berlin data set, the ground truth labels are sparse, so 
we extract small image patches as training and testing samples, 
with the center of each image patch aligned with one label. 
Hence LULC classification on Berlin data set becomes a 
image classification task. 
We select ResNet18 and ResNet50 as baseline, and experiment 
with SepG-ResNet18 and SepG-ResNet50, where regular convolution 
layers in the baseline models are replaced with SepDGConv layers. 
In addition, we compare with S2FL\cite{hong2021multimodal}, 
which achieves SOTA performance on this data set. 

\textbf{Implementation details} For SepG-ResNet18,  
ResNet18, SepG-ResNet50 and ResNet50,
we use image patch of $17 \times 17$ as training 
and test samples. 

To train SepG-ResNet18 and ResNet18 we use stochastic 
gradient descent with momentum as our optimizer, with 
momentum parameter set to 0.9. We train the networks for 
300 epochs. Initial learning rate is 
set to 0.001. Both 
models are trained on a single GPU using a batch size of 
64. For SepG-ResNet50 and ResNet50, we use Adam as 
optimizer, with default algorithm 
parameters. We train both networks for 
400 epochs. Initial learning rate is 
set to 0.001, which is further decayed to 0.0001 at the 
300th epoch. 
Both models are trained on a single GPU using a batch size of 
64. 

\textbf{Results} The quantitative classification results of 
SepG-ResNet18, ResNet18, SepG-ResNet50, ResNet50 and 
S2FL are shown in 
Table V, while Fig. 6 shows the classification 
map of our models, alongside with 
the ground truth labels. In Table V, we directly cite 
the results of S2FL from 
\cite{hong2021multimodal}. While S2FL is not DL-based, it 
is essentially multi-stream and 
achieves SOTA performance on Berlin data set so we believe 
it is worthy of reference. Our 
experimental results show that ResNet based methods generally 
outperform S2FL. SepG-ResNet18 surpasses its baseline 
model and improves SOTA OA on Berlin data set 
to $68.21\%$, while SepG-ResNet50 obtains a marginally lower 
classification accuracy than basic ResNet50. We will 
see in Sec. IV(C) that the best performance is achieved using 
the ResNet50 architecture when some but not all convolution 
layers are replaced with SepDGConv. Variance reduction is also 
observed on SepDGConv models. Fig. 6 shares a similar 
visual comparison with quantitative results.

SepDGConv group structure plots for SepG-ResNet18 and 
SepG-ResNet50 are shown in Fig. 8(b) and 8(c). 
The \textit{Sparsity} plot shows that both learned 
architectures are generally sparse, however sparsity drop 
in shallow layers, which suggests early fusion, is also 
found present in SepG-ResNet18 and SepG-ResNet50 on 
Berlin data set. The \textit{\#Groups} plot shows that, 
in SepG-ResNet18, InConv learns $\sim 10$ groups, while 
in SepG-ResNet50, InConv learns $2 - 4$ groups, which is 
in contrary to the empirical principle that the number 
of groups should be the same as the number of sensors. 
Besides, there are generally more groups in the last 
two blocks than in previous blocks for both models. Since 
in ResNets there are more feature maps and thus more parameters 
in Layer3 and Layer4, this result is also consistent 
with SepG-UNet. 

%------------------------------------------------

\begin{table*}[p]
	\centering
	\caption{Model performance on Houston2018 data set}
	  \begin{tabular}{cr|c|c|c|c}
		\toprule
		\multirow{2}[4]{*}{\# } & \multicolumn{1}{c|}{\multirow{2}[4]{*}{Class}} & \multicolumn{4}{c}{Performance (\%)} \\
	\cmidrule{3-6}          &       & DCNN  & Fusion-FCN & UNet  & SepG-UNet \\
		\midrule
		1     & \multicolumn{1}{c|}{Healthy Grass} & 99.2  & 88.70 & 83.67±1.47 & 82.68±2.61 \\
		2     & \multicolumn{1}{c|}{Stressed Grass} & 0.0   & 87.30 & 67.17±2.65 & 69.11±2.43 \\
		3     & \multicolumn{1}{c|}{Artificial Turf} & 21.3  & 64.14 & 39.85±17.13 & 56.46±22.66 \\
		4     & \multicolumn{1}{c|}{Evergreen Trees} & 95.6  & 97.05 & 82.20±2.15 & 84.00±1.31 \\
		5     & \multicolumn{1}{c|}{Deciduous Trees} & 59.7  & 73.02 & 63.25±4.41 & 67.78±7.57 \\
		6     & \multicolumn{1}{c|}{Bare Earth} & 5.4   & 27.64 & 52.05±4.43 & 48.05±4.51 \\
		7     & \multicolumn{1}{c|}{Water} & 96.7  & 9.15  & 24.43±20.68 & 16.83±1.51 \\
		8     & \multicolumn{1}{c|}{Residential Buildings} & 0.0   & 75.03 & 71.72±4.66 & 70.50±4.62 \\
		9     & \multicolumn{1}{c|}{Non-residential Buildings} & 95.1  & 93.55 & 64.73±2.63 & 61.87±1.63 \\
		10    & \multicolumn{1}{c|}{Roads} & 76.0  & 62.44 & 52.25±1.79 & 52.13±3.25 \\
		11    & \multicolumn{1}{c|}{Sidewalks} & 69.5  & 68.52 & 62.68±3.33 & 63.56±4.12 \\
		12    & \multicolumn{1}{c|}{Crosswalks} & 0.0   & 7.46  & 26.35±4.35 & 38.02±13.80 \\
		13    & \multicolumn{1}{c|}{Major Thoroughfares} & 33.3  & 59.94 & 35.38±0.81 & 31.53±0.70 \\
		14    & \multicolumn{1}{c|}{Highways} & 30.3  & 17.95 & 43.94±6.26 & 33.16±4.79 \\
		15    & \multicolumn{1}{c|}{Railways} & 85.4  & 80.46 & 36.23±14.58 & 48.44±10.28 \\
		16    & \multicolumn{1}{c|}{Paved Parking Lots} & 58.4  & 60.80 & 71.44±3.46 & 72.55±8.09 \\
		17    & \multicolumn{1}{c|}{Unpaved Parking Lots} & 0.0   & 0.00  & 0.00±0.00   & 0.00±0.00 \\
		18    & \multicolumn{1}{c|}{Cars} & 0.0   & 64.33 & 71.45±2.36 & 74.87±3.96 \\
		19    & \multicolumn{1}{c|}{Trains} & 89.6  & 50.94 & 92.44±1.36 & 88.30±5.93 \\
		20    & \multicolumn{1}{c|}{Stadium Seats} & 85.1  & 41.97 & 75.46±9.16 & 71.06±8.49 \\
		\midrule
		\multicolumn{2}{c|}{AA} & 50.0  & 56.52 & 55.84±5.38 & \textbf{56.55±5.61} \\
		\multicolumn{2}{c|}{OA} & 51.2  & 63.28 & 63.66±1.39 & \textbf{63.74±1.05} \\
		\multicolumn{2}{c|}{Kappa} & 0.48  & 0.61  & 0.61±0.01 & \textbf{0.62±0.01} \\
		\bottomrule
	  \end{tabular}%
	\label{tab:result_houston}%
\end{table*}%

\begin{table*}[p]
	\centering
	\caption{Model performance on Berlin data set}
	  \begin{tabular}{cr|c|c|c|c|c}
	  \toprule
	  \multirow{2}[4]{*}{\# } & \multicolumn{1}{c|}{\multirow{2}[4]{*}{Class}} & \multicolumn{5}{c}{Performance (\%)} \\
  \cmidrule{3-7}          &       & S2FL  & ResNet18 & SepG-ResNet18 & ResNet50 & SepG-ResNet50 \\
	  \midrule
	  1     & \multicolumn{1}{c|}{Forest} & 83.30 & 61.03±7.67 & 68.25±2.59 & 63.74±3.35 & 65.73±3.87 \\
	  2     & \multicolumn{1}{c|}{Residential Area} & 57.39 & 76.92±4.26 & 79.85±2.52 & 78.73±3.16 & 78.28±1.16 \\
	  3     & \multicolumn{1}{c|}{Industrial Area} & 48.53 & 37.43±2.47 & 39.26±3.06 & 38.78±3.11 & 34.65±3.40 \\
	  4     & \multicolumn{1}{c|}{Low Plants} & 77.16 & 75.57±2.75 & 74.70±1.63 & 75.85±1.64 & 73.62±3.02 \\
	  5     & \multicolumn{1}{c|}{Soil} & 83.84 & 69.60±3.91 & 68.26±4.68 & 66.02±3.10 & 68.63±7.56 \\
	  6     & \multicolumn{1}{c|}{Allotment} & 57.05 & 19.64±2.56 & 25.47±2.16 & 23.21±2.07 & 27.51±4.32 \\
	  7     & \multicolumn{1}{c|}{Commercial Area} & 31.02 & 27.52±1.03 & 25.02±3.33 & 31.60±3.68 & 26.89±3.07 \\
	  8     & \multicolumn{1}{c|}{Water} & 61.57 & 60.04±1.73 & 53.05±1.95 & 58.05±5.47 & 57.71±8.17 \\
	  \midrule
	  \multicolumn{2}{c|}{AA} & \textbf{62.23} & 53.47±3.30 & 54.23±2.74 & 54.49±3.20 & 54.13±4.32 \\
	  \multicolumn{2}{c|}{OA} & 62.48 & 65.43±3.56 & \textbf{68.21±2.43} & 66.98±2.73 & 66.32±0.72 \\
	  \multicolumn{2}{c|}{Kappa} & 0.49  & 0.51±0.04 & \textbf{0.54±0.03} & 0.53±0.03 & 0.53±0.01 \\
	  \bottomrule
	  \end{tabular}%
	\label{tab:result_berlin}%
\end{table*}%

\begin{table*}[p]
	\centering
	\caption{Model performance on MUUFL data set}
	% \resizebox{2\columnwidth}{!}
	% {
	  \begin{tabular}{cr|c|c|c|c|c|c}
		\toprule
		\multirow{2}[4]{*}{\# } & \multicolumn{1}{c|}{\multirow{2}[4]{*}{Class}} & \multicolumn{6}{c}{Performance (\%)} \\
	\cmidrule{3-8}          &       & OTVCA & TB-CNN & ResNet18 & SepG-ResNet18 & ResNet50 & SepG-ResNet50 \\
		\midrule
		1     & \multicolumn{1}{c|}{Trees} & 84.74±7.28 & 89.97±1.12 & 93.33±0.43 & 92.13±1.11 & 92.23±1.68 & 91.81±0.47 \\
		2     & \multicolumn{1}{c|}{Mostly Grass} & 82.47±0.71 & 80.61±4.98 & 72.76±1.74 & 71.82±1.91 & 71.96±3.95 & 72.90±2.72 \\
		3     & \multicolumn{1}{c|}{Mixed Ground Surface} & 69.93±1.40 & 73.25±8.15 & 75.91±1.18 & 74.56±1.60 & 73.71±1.55 & 73.55±1.89 \\
		4     & \multicolumn{1}{c|}{Dirt and Sand} & 85.93±0.62 & 83.46±1.48 & 85.16±1.11 & 85.73±1.54 & 83.95±1.60 & 83.88±1.22 \\
		5     & \multicolumn{1}{c|}{Road} & 83.76±2.84 & 88.04±0.48 & 90.25±0.75 & 87.93±2.12 & 86.9±1.09 & 83.33±1.57 \\
		6     & \multicolumn{1}{c|}{Water} & 99.45±1.14 & 67.60±0.13 & 73.74±6.60 & 76.23±7.62 & 74.73±4.94 & 78.73±6.07 \\
		7     & \multicolumn{1}{c|}{Building Shadow} & 91.19±0.61 & 84.55±3.01 & 79.25±2.19 & 75.73±3.30 & 82.33±1.78 & 80.61±2.77 \\
		8     & \multicolumn{1}{c|}{Building} & 96.66±5.85 & 92.92±0.81 & 95.26±0.41 & 91.39±3.17 & 94.11±0.66 & 88.94±0.74 \\
		9     & \multicolumn{1}{c|}{Sidewalk} & 75.80±0.39 & 67.73±4.46 & 65.78±2.24 & 63.40±3.15 & 54.17±2.79 & 56.68±3.00 \\
		10    & \multicolumn{1}{c|}{Yellow Curb} & 89.16±0.33 & 17.49±2.71 & 31.10±4.13 & 30.80±6.28 & 22.54±2.94 & 24.14±3.16 \\
		11    & \multicolumn{1}{c|}{Cloth Panels} & 98.22±0.31 & 43.12±0.41 & 83.17±4.67 & 82.97±2.78 & 70.69±6.19 & 65.62±7.58 \\
		\midrule
		\multicolumn{2}{c|}{AA} & 84.15±0.56 & \textbf{85.47±1.05} & 76.88±2.31 & 75.70±3.14 & 73.39±2.65 & 72.75±2.84 \\
		\multicolumn{2}{c|}{OA} & 79.57±2.00 & 84.47±1.27 & \textbf{86.44±0.11} & 84.6±1.85 & 84.28±1.28 & 83.23±0.65 \\
		\multicolumn{2}{c|}{Kappa} & \textbf{0.87±0.01} & 0.72±0.02 & 0.83±0.00 & 0.80±0.02 & 0.80±0.02 & 0.79±0.01 \\
		\bottomrule
	  \end{tabular}%
	% }
	\label{tab:result_muufl}%
\end{table*}%

%------------------------------------------------
\vspace{6pt}

\subsubsection{Experiments on MUUFL data set} \quad

\begin{figure*}[htb!]
	\begin{center}
			\includegraphics[width=2\columnwidth]{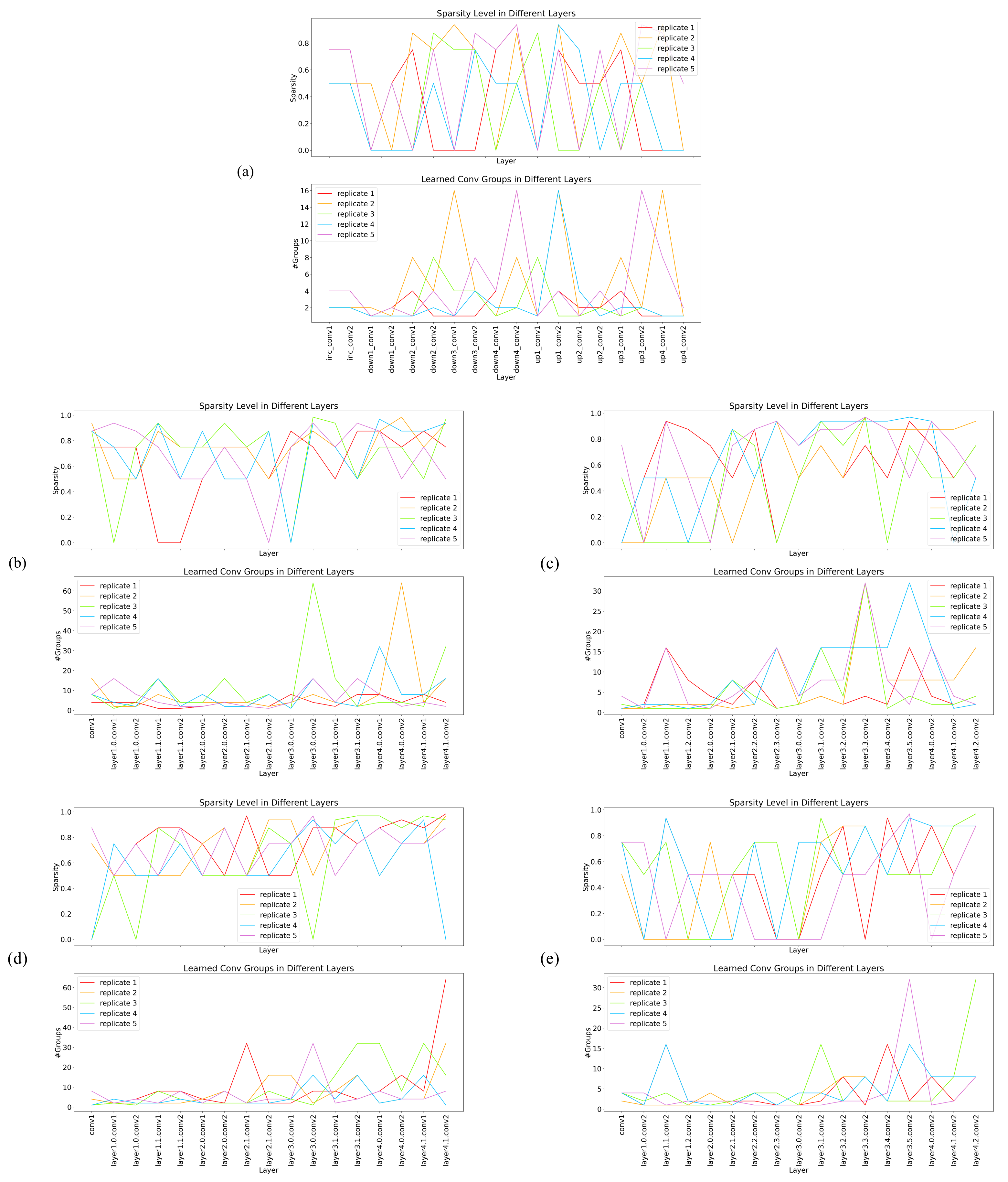}
		\caption{Number of 
		groups and sparsity of U in each layer of 
		the following SepDGConv-based models on 
		different data sets: 
		(a) SepG-UNet trained on Houston2018 data set, 
		(b) SepG-ResNet18 trained on Berlin data set, 
		(c) SepG-ResNet50 trained on Berlin data set, 
		(d) SepG-ResNet18 trained on MUUFL data set, and
		(e) SepG-ResNet50 trained on MUUFL data set, 
		}
	\label{fig:groups}
	\end{center}
\end{figure*}

For MUUFL data set we follow most studies on it and use 
classification models. We experiment with ResNet18, ResNet50 
and their Sep-DGConv derivatives on MUUFL data set. We also 
compare our results with those of the following methods: 
orthogonal total variation component analysis (OTVCA) 
\cite{rasti2017hyperspectral}, and two branch CNN (TB-CNN) 
\cite{xu2017multisource}. Both OTVCA and TB-CNN are 
multi-stream models. 

\textbf{Implementation details} For SepG-ResNet18 and 
ResNet18, we use image patch of $11 \times 11$ as training 
and test samples, while for SepG-ResNet50 and ResNet50, 
we use image patch of size $17 \times 17$. Since previous 
studies do not use a fixed training set, we make random 
train-test split in each replica under different 
random seeds. We follow \cite{zhao2021fractional} 
and set training set size 
fixed to 100 and the rest used as test set. 

For SepG-ResNet18 and 
ResNet18, we use He initialization \cite{he2015delving}
to initialize convolution filters, and zero initialization 
for the last batch normalization layer in each residual branch
\cite{goyal2017accurate}. To train the networks we use 
stochastic 
gradient descent with momentum as our optimizer, with 
momentum parameter set to 0.9. We train the networks for 
300 epochs. Initial learning rate is 
set to 0.02, with a learning rate schedule that decreases 
the learning rate to 0.002 at 200th epoch, and further 
decreases the learning rate to 0.0002 at 240th epoch. Both 
models are trained on a single GPU using a batch size of 
48. In SepG-ResNet50 and ResNet50, He initialization 
and zero batch norm 
initialization are also used. 
We use Adam as optimizer, with default algorithm 
parameters. We train both networks for 
400 epochs. Initial learning rate is 
set to 0.01, with a learning rate schedule at epochs 
[300, 350], decreasing the learning rate to [0.001, 0.0001]. 
Both models are trained on a single GPU using a batch size of 
64.

\textbf{Results} The quantitative classification results of 
SepG-ResNet18, SepG-ResNet50, ResNet18, ResNet50, and 
methods to compare 
with are shown in Table VI. These results are cited from 
\cite{zhao2021fractional} who replicated these models 
in their work, since we follow their experimental settings. 
Fig. 7 shows the classification 
map of these models alongside with 
the ground truth labels. According to Table VI, the ResNets 
generally achieve better performance than OTVCA and TB-CNN. 
On MUUFL 
data set, however, both SepG-ResNet18 and SepG-ResNet50 
do not surpass their corresponding baseline model. Again we 
will see in Sec. IV(C) that by using SepDGConv in some but not 
all convolution layers both models can obtain better 
performance and outperform baseline models. In Fig. 7, 
the output classification maps of ResNet50s are generally 
more noisy than those of ResNet18s, which suggest that 
there is probably overfitting in ResNet50s, since the 
data set is relatively small. 

SepDGConv group structures learned on MUUFL data set are 
shown in Fig. 8(d) and 8(e). Both early fusion and more 
groups in deeper layers are consistently observed. 

%------------------------------------------------

\subsection{Ablation analysis}

\begin{figure*}[htb!]
	\begin{center}
			\includegraphics[width=2\columnwidth]{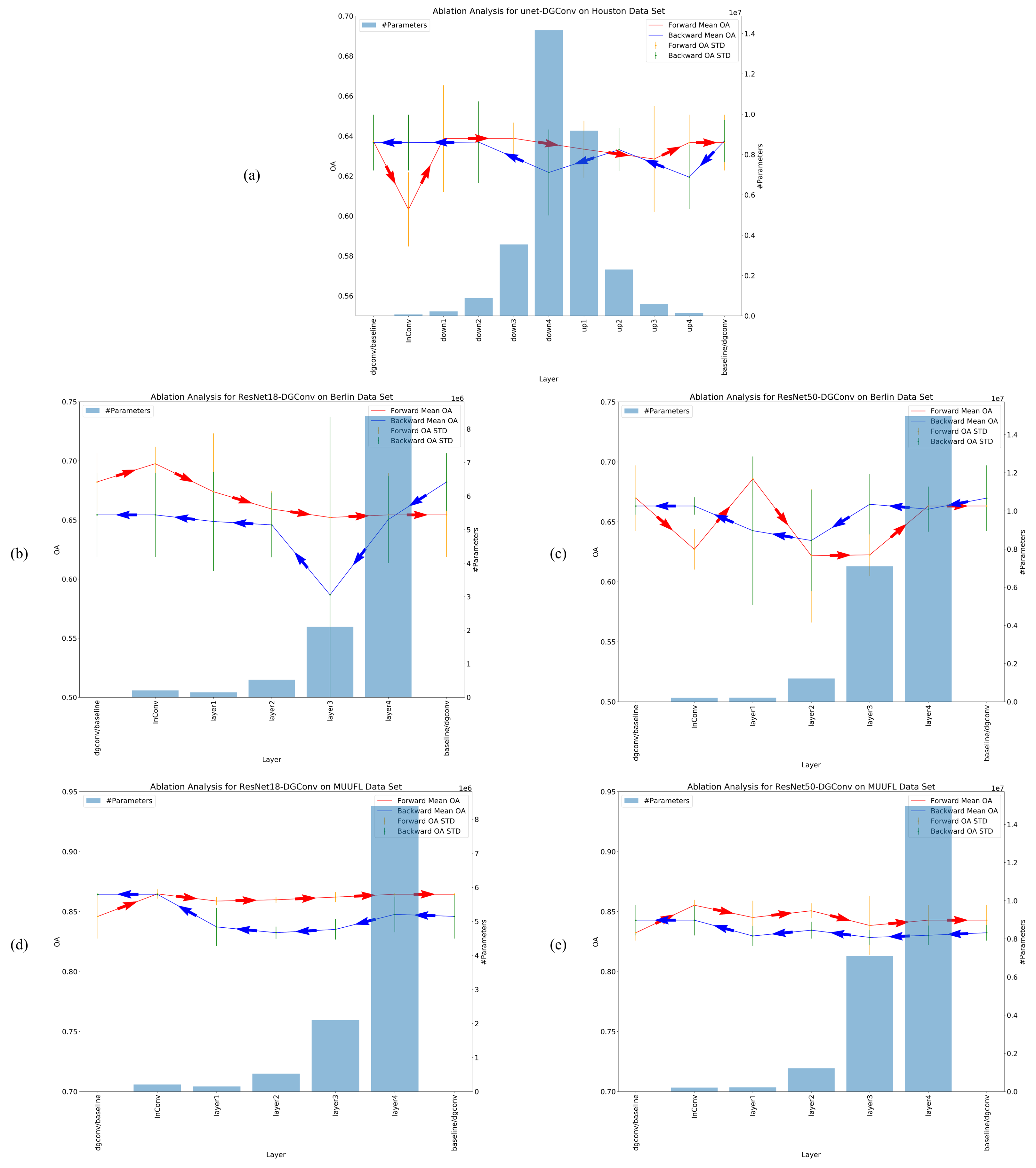}
		\caption{Classification performance of various 
		models obtained via ablation analysis. The arrows 
		indicate the order of removal of SepDGConv modules, while 
		blue bars represent number of parameters in the 
		corresponding block. 
		(a) Performance of SepG-UNets on Houston2018 data set, 
		(b) performance of SepG-ResNet18s on Berlin data set, 
		(c) performance of SepG-ResNet50s on Berlin data set, 
		(d) performance of SepG-ResNet18s on MUUFL data set, and
		(e) performance of SepG-ResNet50s on MUUFL data set, 
		}
	\label{fig:retrain}
	\end{center}
\end{figure*}

To investigate the performance improvement of SepDGConv, we 
remove SepDGConv block-by-block from above models, and analyze 
the influence of SepDGConv's usage on the overall performance. 
In particular, since separate convolution groups learned in a 
SepG-Net represents sensor-specific branches, we hope to shed 
light on whether such sensor-specificity is important for 
multi-source RS data fusion. 

We design the ablation experiments based on the optimal 
brain damage (OBD) theory in DL \cite{lecun1990optimal}, 
according to which if an important module in a deep neural network 
is removed, then significant performance drop should be 
observed. Concretely, 
for each SepDGConv model we run one forward pass and one backward 
pass. In the forward pass, we in order change SepDGConv in the 
following blocks back to regular convolution: [InConv, 
Layer1, Layer2, Layer3, Layer4] for SepG-ResNet, and 
[InConv, Down1, Down2, Down3, Down4, Up1, Up2, Up3, Up4] 
for SepG-UNet. In the backward pass, the 
order is reversed. The obtained models are retrained. 
For each model, the baseline, SepDGConv derivative, forward 
pass and backward pass are trained under one same configuration 
of hyperparameters. 

The results of ablation experiments are shown in Fig. 9. 
We run each model 5 times with random seeds 42-46, same as 
our main experiment above. 
The value of each data point in Fig. 9 is taken as average 
OA of the 5 replicas, while each error bar represents standard 
deviation of 5 OA values. 
The red line represents OA changes in a forward pass, while 
the blue line represents OA changes in a backward pass. A 
data point in a forward pass means that all SepDGConvs in blocks 
previous to and including this point are replaced with 
regular convolution. For example, in Fig. 9(b) a point at 
Layer1 represents results from ResNet18s that have regular 
convolution in InConv and Layer1, and SepDGConv in Layer2-4. 
Similarly, a data point in a backward pass means that all 
SepDGConvs in blocks behind and including this point are replaced 
with regular convolution, while blocks previous to this 
point remain using SepDGConv. 

In forward passes, performance gain is consistently observed 
in early stage of each pass. For example, in Fig. 9(b), 
on Berlin data set SepG-ResNet18's performance improves 
as SepDGConv is removed from InConv and an OA of $69.76\%$ is 
obtained, which surpasses both the original SepG-Net 
as well as the baseline. Same performance gain can be 
observed in Fig. 9(d) and 9(e), when SepDGConv in InConv is 
replaced by regular convolution. According to OBD theory, 
this phenomenon implies that, rather than being beneficial 
to model performance, imposing multi-stream architecture 
in shallow layers actually \textit{harms} model performance. 
For SepG-ResNet50 on 
Berlin data set, as shown in Fig. 9(c), while there is an 
initial performance loss, the OA goes up to $68.58\%$ as 
soon as SepDGConv layers in both InConv and Layer1 are removed, 
which means that at this point the shallow layers in the 
model are densely connected rather than having a multi-stream, 
sensor-specific
architecture. A very similar loss-gain curve is observed 
on SepG-UNet, as shown in Fig. 9(a). 
Hence experimental results of SepG-ResNet50 on Berlin 
data set and SepG-UNet on Houston2018 data set both 
support our finding that for the first few blocks dense 
convolution is better than multi-stream convolution. 

In backward passes, performance loss is observed 
as SepDGConv in middle and deep layers are removed. 
For UNet, the performance loss occurs as the backward 
pass goes through middle blocks, from Up1 to Down4, 
as shown in Fig. 9(a). For 
ResNets, we observe OA drop when SepDGConv in the 
last two blocks, Layer4 and Layer3, is replaced, as 
shown in Fig 9(b)-(d). 

Furthermore, based on the distribution of number of 
parameters in the studied models, shown as the blue bars 
in Fig. 9, we summarize the performance 
loss in middle and last layers into a more general 
phenomenon, i.e., performance loss in wide layers. Wide 
layers refer to layers with more feature map channels, and 
thus with more parameters. It can be seen from Fig. 9 that 
for UNet the wide layers are Down3-Up2, exactly where we 
observe performance loss, and for ResNets the wide layers 
are Layer3-Layer4, where performance loss in the backward 
pass is also observed. Such performance loss implies that 
SepDGConv is beneficial to model performance if applied to 
wide layers. 

Finally we shall revisit the performance gain phenomenon 
observed in the early stage of forward passes. This phenomenon 
occurs in the first few layers of studied models, where 
there are fewer parameters, as shown in Fig. 9. These layers 
are also called narrow layers, in accordance with "wide layers". 
Thus we can say that there is performance gain if we replace 
SepDGConv with regular convolution in narrow layers. Such 
performance gain implies that SepDGConv is harmful to model 
performance if applied to narrow, usually shallow layers. 
Besides, in Fig. 9(b)-(e), performance gain is observed 
in backward passes going from Layer2 to InConv. This is also 
a clue that dense convolution in narrow layers is 
more favorable to model performance. 

To summarize, in the ablation analysis we observe that there 
is model performance gain if we replace SepDGConv with regular 
convolution in narrow, usually the first few layers in a 
model, and there is model performance loss if we replace 
SepDGConv with regular convolution in wide, usually the last 
few layers in a model. These findings imply that multi-stream 
architecture is harmful to model performance if used in 
narrow layers, but becomes beneficial if applied to wide 
layers. 

%------------------------------------------------

\subsection{Comparing different convolution strategies}

\begin{table*}[hb!]
	\centering
	\caption{Different Convolution Strategies on UNet, Houston2018 Data Set}
		\begin{tabular}{c|cccccc}
		\toprule
				& UNet  & GConv & FGConv & SepDGConv & Ablation & False Group \\
		\midrule
		OA (\%) & 63.66±1.89 & 65.14±2.52 & 65.49±0.62 & 63.74±1.05 & 63.88±0.79 & 65.33±1.51 \\
		\bottomrule
		\end{tabular}%
	\label{tab:group_unet}%
\end{table*}%

Results of our ablation analysis suggest that it is better 
to use regular convolution than SepDGConv in the first 
few layers of a CNN. This indicates that may not be playing 
an important role in data fusion models, since by using 
densely connected convolution the model no longer has 
sensor-specific features. However, SepDGConv also regularizes 
the model and the effect of sensor-specificity is not 
yet isolated from regularization. In this subsection, 
we further compare models using SepDGConv with another 
two convolution strategies that impose sensor-specific 
multi-branch architectures: group convolution (GConv) 
and fixed groupable convolution (FGConv). 

As mentioned above (Sec. II(A)), by using GConv for the 
first $l$ layers and setting the number of groups to a 
fixed number $G_0$, a CNN with $G_0$ sensor-specific 
branches, each of depth $l$, can be built. In our experiments 
below we will use this strategy to construct models 
that have strictly sensor-specific branches, so we can 
better observe the effect of sensor-specificity. 
Specifically, for ResNets we will use GConv for 
[InConv, Layer 1-4], and for UNets we will use GConv 
for [InConv, Down1-4, Up1-4]. 

The role of FGConv is to further isolate sensor-specificity 
from regularization. Using SepDGConv does not guarantee 
sensor-specificity of deep layers, since the number of 
groups it learns varies from layer to layer. The idea of 
FGConv is to combine SepDGConv and GConv to make sure 
that any layer using FGConv has at least $G_0$ groups. 
Specifically, GConv can be equivalently expressed by an 
relationship matrix $U_0$ (Fig. 2(c)), and in FGConv we 
construct a new relationship matrix $U_F$ for a layer, 
using that layer's learned SepDGConv relationship matrix 
$U$ and a specified GConv matrix $U_0$: 

\begin{equation}\label{equ:16}
	U_F = U_0 \odot U
\end{equation}
where $\odot$ denotes element-wise product. $U_0$ is 
constructed using the pre-specified group number 
$G_0$ and is fixed, while $U$ is still learned from 
data. In the experiments, FGConv is used in 
[InConv, Layer 1-4] for ResNets, and 
[InConv, Down1-4, Up1-4] for UNets. 

For GConv, we make sure at the input layer that the group 
division leads to sensor-specificity, by manually designing 
the relationship matrix $U_0$. For Houston2018 data set, 
$U_0$ divides the 58 channel input data into 4 groups, 
mapping [3 MS, 48 HS, 3 LiDAR, 4 DEM/DSM] to [16, 16, 16, 
16] feature maps. For Berlin data set, $U_0$ maps 
[244 HS, 4 SAR] to [32, 32] feature maps. For MUUFL 
data set, $U_0$ maps [64 HS, 2 LiDAR] to [32, 32] 
feature maps. The same $U_0$ is applied to FGConv. 

Experimental results are shown in Table VII-XI. Performance 
of baseline models and SepG-models are cited from above, 
while the "Ablation" column refers to the model's best 
performance obtained in the ablation study. We report 
average and std of OA of 5 runs, using random seeds 42-46, 
same as above. 

\begin{table}[htbp]
	\centering
	\caption{Different Convolution Strategies on ResNet18, Berlin Data Set}
	\resizebox{\columnwidth}{!}
	{
		\begin{tabular}{c|ccccl}
		\toprule
				& ResNet18 & GConv & FGConv & SepDGConv & \multicolumn{1}{c}{Ablation} \\
		\midrule
		OA (\%) & 65.43±3.56 & 65.04±2.90 & 68.09±3.94 & 68.21±2.43 & 69.76±1.43 \\
		\bottomrule
		\end{tabular}%
	}
	\label{tab:group_berlin18}%
\end{table}%

\begin{table}[htbp]
	\centering
	\caption{Different Convolution Strategies on ResNet50, Berlin Data Set}
	\resizebox{\columnwidth}{!}
	{
		\begin{tabular}{c|ccccc}
		\toprule
				& ResNet50 & GConv & FGConv & SepDGConv & Ablation \\
		\midrule
		OA (\%) & 66.98±2.73 & 65.90±1.24 & 64.85±3.07 & 66.32±0.72 & 68.58±1.57 \\
		\bottomrule
		\end{tabular}%
	}
	\label{tab:group_berlin50}%
\end{table}%

\begin{table}[htbp]
	\centering
	\caption{Different Convolution Strategies on ResNet18, MUUFL Data Set}
	\resizebox{\columnwidth}{!}
	{
		\begin{tabular}{c|ccccc}
		\toprule
				& ResNet18 & GConv & FGConv & SepDGConv & Ablation \\
		\midrule
		OA (\%) & 86.44±0.11 & 86.17±0.32 & 86.40±0.53 & 84.60±1.85 & 86.47±0.39 \\
		\bottomrule
		\end{tabular}%
	}
	\label{tab:group_muufl18}%
\end{table}%

\begin{table}[htbp]
	\centering
	\caption{Different Convolution Strategies on ResNet50, MUUFL Data Set}
	\resizebox{\columnwidth}{!}
	{
		\begin{tabular}{c|ccccc}
		\toprule
				& ResNet50 & GConv & FGConv & SepDGConv & Ablation \\
		\midrule
		OA (\%) & 84.28±1.28 & 83.60±1.03 & 85.30±0.20 & 83.23±0.65 & 85.53±0.45 \\
		\bottomrule
		\end{tabular}%
	}
	\label{tab:group_muufl50}%
\end{table}%

First consider the ResNets, Table VIII-XI. 
(1) It is consistently observed 
that baseline models marginally outperform the 
corresponding GConv models. This indicates that 
sensor-specific multi-branching does not 
necessarily help improve model performance. 
(2) In 3 out of total 4 experiments, i.e., 
except ResNet50 on Berlin data set, FGConv 
outperforms GConv. This supports our 
basic assumption that automatically learning 
group convolution hyperparameters benefits 
model performance. (3) In all of 4 experiments, 
neither does GConv nor does FGConv outperform 
the best architecture we previously find in the 
ablation study, where the models' first few 
SepDGConv layers are relpaced with regular 
convolution. These models found in the ablation 
study do not have sensor-specific branches, so 
this result is consistent with (1). 

Then consider the UNets on Houston2018 data 
set, Table VII. While in consistent with (2) 
above that FGConv outperforms GConv, for UNets 
we observe GConv outperforming baseline, SepDGConv 
and Ablation. To find an explanation for this 
we add a False-GConv experiment, where at the 
input layer we use a group division of 
[14, 16, 14, 14] instead of [2, 48, 3, 4], so 
that the 4 branches are no longer sensor-specific. 
False-GConv outperforms GConv while achieves 
slightly lower accuracy than FGConv. This implies 
that GConv and FGConv gains performance probably 
from the setting $G=4$, and again this experiment 
supports our finding above that sensor-specificity 
is not necessarily helpful. 

%------------------------------------------------

\subsection{Discussion}

\subsubsection{Multi-stream as regularization} 
Theoretically, 
both SepDGConv and human designed multi-stream deep neural 
networks can be regarded as regularization technique 
since it imposes certain constraints on the network 
architecture in order to reduce overfitting and to
improve model performance 
\cite{goodfellow2016deep} \cite{kukavcka2017regularization}. 
Based on our experiments and ablation analysis above, 
we attribute model performance in multi-source RS data 
fusion by using SepDGConv to regularization, for we have 
observed the following two very important features of 
regularization: 

(1) Variance reduction. It is known that regularized models 
can generalize better, which means they should have less 
test variance. In our experiments, as discussed in 
Sec. III(B), except ResNet18 on MUUFL data set, all SepDGConv 
models have less test OA variance than their corresponding 
baseline models. 

(2) Over-regularization. If a simple model is regularized 
too much, the model capacity can be reduced too much to 
fit the data, and as a result the overall model performance 
is harmed. In our ablation analysis, as shown in Fig. 9, 
we find out that imposing multi-stream SepDGConv on shallow 
layers leads to model performance loss. The most possible 
reason is that, these shallow, narrow layer themselves 
do not have many parameters, and by using SepDGConv they are 
over-regularized, leading to underfitting. On the other hand, 
the middle layers in UNet and last few layers in ResNets 
are much wider and have much more parameters than the 
first few layers, hence using SepDGConv on these wide layers 
very likely 
just achieves the expected regularization effect, improving 
the models' performance. 

Since we experiment with 3 different models on 3 very 
diverse data sets, our two findings above are highly 
generalizable, which provides strong clues that multi-stream 
architectures actually play the role of model regularizer. 

Yet, regularization itself is a complex technique and its 
effect is always coupled with various aspects of model 
optimization and generalization, hence there are probably 
many other factors to explore that contribute to the 
phenomena we have observed. For example,  
a very recent paper \cite{zheng2021deep}
find out that one same model trained separately on 
different source of data acquired in the same area 
(RGB and SAR in their case) can lead to very similar model 
parameter distributions. This finding suggests that 
there could be feature redundancy in shallow layers of 
multi-stream architectures. Nevertheless, we hope our 
work shed 
some light on the mechanisms behind neural network 
architecture designing for multi-source RS data, and 
inspire novel research. 

\subsubsection{Possible improvements on group convolution}
Our results suggest that models with regular convolution 
such as ResNet18 can obtain classification results at least 
comparable with SOTA methods, and that in shallow layers 
dense, regular convolution should be used, 
which together advocate single-stream 
deep CNN models for joint classification of multi-source 
RS data. To automatically learn grouped convolution in 
wide layers to utilize the regularization effect, it is 
more desirable if SepDGConv can learn dense convolution for 
narrow layers, and thus there is still room for improvement 
in SepDGConv. Besides, the restrictions SepDGConv puts on the 
relationship matrix $U$ are strong and in practice we may 
need to construct $U$s with more flexible structure. We hope 
novel research in constructing and learning the relationship 
matrix $U$ can lead to better single-stream CNN 
architectures for multi-source RS data. 

\subsubsection{Towards better performance}
Our work makes it possible to build deep, single stream 
networks for multi-source RS data. On the one hand, 
modern techniques that boost model performance are more 
easily applied in a unified network. On the other hand, 
designing sufficiently large models is beneficial to, 
and probably necessary for, solving large-scale, real 
world RS problems. 

In our paper, we only experiment and compare with basic 
models, leaving many techniques such as 
ensembling, and more complex modules such as attention 
mechanism to further improve model performance unexplored. 
For example, \cite{zhao2021fractional} utilizes Octave 
convolution and fractional Gabor convolution to propose 
a network that achieves SOTA OA of $89.90\%$ on MUUFL data 
set. We believe that such advanced modules, and more in 
the future, can be more easily implemented in a 
single-branch network. 

Furthermore, as we can build models of comparable depth to 
those in modern CV research with our proposed 
technique, we can better utilize the benefits of 
overparametrized models, as well as potentially available 
large RS data in the future. 

%------------------------------------------------

\section{CONCLUSION}\label{sec:CONCLUSION}

In this paper we have investigated the potential of 
single-stream models in joint classification of multi-source 
remote sensing data. To enable multi-stream network structure 
to be automatically learned within a single-stream 
architecture, we propose the SepDGConv module based on 
group 
convolution and dynamic grouping convolution technique. 
With reference to modern deep convolutional 
neural network architectures, we then propose several deep 
learning models with SepDGConv: SepG-ResNet18, SepG-ResNet50, 
and SepG-UNet. The propsoed models are verified on three 
benchmark data sets with diverse data modality, yielding 
promising classification results, which indicate the 
effectiveness and generalizability 
of proposed single-stream networks for 
multi-source remote sensing data joint classification. 
Furthermore, we analyze the usage of SepDGConv in different 
parts of the models and find out that: (1) using SepDGConv generally 
reduces model variance, (2) using SepDGConv in narrow layers, 
usually the first few layers, 
harms model performance, and (3) using SepDGConv in wide 
layers, usually the last few layers, 
improves model performance. These findings imply 
that sensor-specific multi-stream architecture is essentially 
playing the 
role of model regularizer, and is not strictly necessary 
for multi-source remote sensing data fusion. We hope our 
work can inspire novel flexible and generalizable models 
for multi-source remote sensing data analysis.

% or


% \appendices

% % you can choose not to have a title for an appendix
% % if you want by leaving the argument blank
% \section{}

% \section*{Acknowledgment}


\ifCLASSOPTIONcaptionsoff
  \newpage
\fi



%\bibliographystyle{IEEEtran}


\bibliographystyle{IEEEtran}
\bibliography{IEEEabrv,references}

\begin{thebibliography}{10}
\providecommand{\url}[1]{#1}
\csname url@samestyle\endcsname
\providecommand{\newblock}{\relax}
\providecommand{\bibinfo}[2]{#2}
\providecommand{\BIBentrySTDinterwordspacing}{\spaceskip=0pt\relax}
\providecommand{\BIBentryALTinterwordstretchfactor}{4}
\providecommand{\BIBentryALTinterwordspacing}{\spaceskip=\fontdimen2\font plus
\BIBentryALTinterwordstretchfactor\fontdimen3\font minus
  \fontdimen4\font\relax}
\providecommand{\BIBforeignlanguage}[2]{{%
\expandafter\ifx\csname l@#1\endcsname\relax
\typeout{** WARNING: IEEEtran.bst: No hyphenation pattern has been}%
\typeout{** loaded for the language `#1'. Using the pattern for}%
\typeout{** the default language instead.}%
\else
\language=\csname l@#1\endcsname
\fi
#2}}
\providecommand{\BIBdecl}{\relax}
\BIBdecl

\bibitem{li2020review}
J.~Li, Y.~Pei, S.~Zhao, R.~Xiao, X.~Sang, and C.~Zhang, ``A review of remote
  sensing for environmental monitoring in china,'' \emph{Remote Sensing},
  vol.~12, no.~7, p. 1130, 2020.

\bibitem{sishodia2020applications}
R.~P. Sishodia, R.~L. Ray, and S.~K. Singh, ``Applications of remote sensing in
  precision agriculture: A review,'' \emph{Remote Sensing}, vol.~12, no.~19, p.
  3136, 2020.

\bibitem{xu2019advanced}
Y.~Xu, B.~Du, L.~Zhang, D.~Cerra, M.~Pato, E.~Carmona, S.~Prasad, N.~Yokoya,
  R.~H{\"a}nsch, and B.~Le~Saux, ``Advanced multi-sensor optical remote sensing
  for urban land use and land cover classification: Outcome of the 2018 ieee
  grss data fusion contest,'' \emph{IEEE Journal of Selected Topics in Applied
  Earth Observations and Remote Sensing}, vol.~12, no.~6, pp. 1709--1724, 2019.

\bibitem{pedergnana2012classification}
M.~Pedergnana, P.~R. Marpu, M.~Dalla~Mura, J.~A. Benediktsson, and L.~Bruzzone,
  ``Classification of remote sensing optical and lidar data using extended
  attribute profiles,'' \emph{IEEE Journal of Selected Topics in Signal
  Processing}, vol.~6, no.~7, pp. 856--865, 2012.

\bibitem{khodadadzadeh2015fusion}
M.~Khodadadzadeh, J.~Li, S.~Prasad, and A.~Plaza, ``Fusion of hyperspectral and
  lidar remote sensing data using multiple feature learning,'' \emph{IEEE
  Journal of Selected Topics in Applied Earth Observations and Remote Sensing},
  vol.~8, no.~6, pp. 2971--2983, 2015.

\bibitem{rasti2017hyperspectral}
B.~Rasti, P.~Ghamisi, and R.~Gloaguen, ``Hyperspectral and lidar fusion using
  extinction profiles and total variation component analysis,'' \emph{IEEE
  Transactions on Geoscience and Remote Sensing}, vol.~55, no.~7, pp.
  3997--4007, 2017.

\bibitem{gu2015novel}
Y.~Gu, Q.~Wang, X.~Jia, and J.~A. Benediktsson, ``A novel mkl model of
  integrating lidar data and msi for urban area classification,'' \emph{IEEE
  transactions on geoscience and remote sensing}, vol.~53, no.~10, pp.
  5312--5326, 2015.

\bibitem{hu2019mima}
J.~Hu, D.~Hong, and X.~X. Zhu, ``Mima: Mapper-induced manifold alignment for
  semi-supervised fusion of optical image and polarimetric sar data,''
  \emph{IEEE Transactions on Geoscience and Remote Sensing}, vol.~57, no.~11,
  pp. 9025--9040, 2019.

\bibitem{singh2007topological}
G.~Singh, F.~M{\'e}moli, G.~E. Carlsson \emph{et~al.}, ``Topological methods
  for the analysis of high dimensional data sets and 3d object recognition.''
  \emph{PBG@ Eurographics}, vol.~2, 2007.

\bibitem{lecun2015deep}
Y.~LeCun, Y.~Bengio, and G.~Hinton, ``Deep learning,'' \emph{nature}, vol. 521,
  no. 7553, pp. 436--444, 2015.

\bibitem{lecun2010convolutional}
Y.~LeCun, K.~Kavukcuoglu, and C.~Farabet, ``Convolutional networks and
  applications in vision,'' in \emph{Proceedings of 2010 IEEE international
  symposium on circuits and systems}.\hskip 1em plus 0.5em minus 0.4em\relax
  IEEE, 2010, pp. 253--256.

\bibitem{marmanis2015deep}
D.~Marmanis, M.~Datcu, T.~Esch, and U.~Stilla, ``Deep learning earth
  observation classification using imagenet pretrained networks,'' \emph{IEEE
  Geoscience and Remote Sensing Letters}, vol.~13, no.~1, pp. 105--109, 2015.

\bibitem{maggiori2016convolutional}
E.~Maggiori, Y.~Tarabalka, G.~Charpiat, and P.~Alliez, ``Convolutional neural
  networks for large-scale remote-sensing image classification,'' \emph{IEEE
  Transactions on geoscience and remote sensing}, vol.~55, no.~2, pp. 645--657,
  2016.

\bibitem{yuan2020review}
X.~Yuan, J.~Shi, and L.~Gu, ``A review of deep learning methods for semantic
  segmentation of remote sensing imagery,'' \emph{Expert Systems with
  Applications}, p. 114417, 2020.

\bibitem{hu2015deep}
W.~Hu, Y.~Huang, L.~Wei, F.~Zhang, and H.~Li, ``Deep convolutional neural
  networks for hyperspectral image classification,'' \emph{Journal of Sensors},
  vol. 2015, 2015.

\bibitem{li2016hyperspectral}
W.~Li, G.~Wu, F.~Zhang, and Q.~Du, ``Hyperspectral image classification using
  deep pixel-pair features,'' \emph{IEEE Transactions on Geoscience and Remote
  Sensing}, vol.~55, no.~2, pp. 844--853, 2016.

\bibitem{lee2017going}
H.~Lee and H.~Kwon, ``Going deeper with contextual cnn for hyperspectral image
  classification,'' \emph{IEEE Transactions on Image Processing}, vol.~26,
  no.~10, pp. 4843--4855, 2017.

\bibitem{he2018lidar}
X.~He, A.~Wang, P.~Ghamisi, G.~Li, and Y.~Chen, ``Lidar data classification
  using spatial transformation and cnn,'' \emph{IEEE Geoscience and Remote
  Sensing Letters}, vol.~16, no.~1, pp. 125--129, 2018.

\bibitem{pan2020land}
S.~Pan, H.~Guan, Y.~Chen, Y.~Yu, W.~N. Gon{\c{c}}alves, J.~M. Junior, and
  J.~Li, ``Land-cover classification of multispectral lidar data using cnn with
  optimized hyper-parameters,'' \emph{ISPRS Journal of Photogrammetry and
  Remote Sensing}, vol. 166, pp. 241--254, 2020.

\bibitem{zhao2016convolutional}
J.~Zhao, W.~Guo, S.~Cui, Z.~Zhang, and W.~Yu, ``Convolutional neural network
  for sar image classification at patch level,'' in \emph{2016 IEEE
  International Geoscience and Remote Sensing Symposium (IGARSS)}.\hskip 1em
  plus 0.5em minus 0.4em\relax IEEE, 2016, pp. 945--948.

\bibitem{ma2018ship}
M.~Ma, J.~Chen, W.~Liu, and W.~Yang, ``Ship classification and detection based
  on cnn using gf-3 sar images,'' \emph{Remote Sensing}, vol.~10, no.~12, p.
  2043, 2018.

\bibitem{chen2017deep}
Y.~Chen, C.~Li, P.~Ghamisi, X.~Jia, and Y.~Gu, ``Deep fusion of remote sensing
  data for accurate classification,'' \emph{IEEE Geoscience and Remote Sensing
  Letters}, vol.~14, no.~8, pp. 1253--1257, 2017.

\bibitem{xu2017multisource}
X.~Xu, W.~Li, Q.~Ran, Q.~Du, L.~Gao, and B.~Zhang, ``Multisource remote sensing
  data classification based on convolutional neural network,'' \emph{IEEE
  Transactions on Geoscience and Remote Sensing}, vol.~56, no.~2, pp. 937--949,
  2017.

\bibitem{xu2018multi}
Y.~Xu, B.~Du, and L.~Zhang, ``Multi-source remote sensing data classification
  via fully convolutional networks and post-classification processing,'' in
  \emph{IGARSS 2018-2018 IEEE International Geoscience and Remote Sensing
  Symposium}.\hskip 1em plus 0.5em minus 0.4em\relax IEEE, 2018, pp.
  3852--3855.

\bibitem{hong2020more}
D.~Hong, L.~Gao, N.~Yokoya, J.~Yao, J.~Chanussot, Q.~Du, and B.~Zhang, ``More
  diverse means better: Multimodal deep learning meets remote-sensing imagery
  classification,'' \emph{IEEE Transactions on Geoscience and Remote Sensing},
  vol.~59, no.~5, pp. 4340--4354, 2020.

\bibitem{hong2020x}
D.~Hong, N.~Yokoya, G.-S. Xia, J.~Chanussot, and X.~X. Zhu, ``X-modalnet: A
  semi-supervised deep cross-modal network for classification of remote sensing
  data,'' \emph{ISPRS Journal of Photogrammetry and Remote Sensing}, vol. 167,
  pp. 12--23, 2020.

\bibitem{zhang2021information}
M.~Zhang, W.~Li, R.~Tao, H.~Li, and Q.~Du, ``Information fusion for
  classification of hyperspectral and lidar data using ip-cnn,'' \emph{IEEE
  Transactions on Geoscience and Remote Sensing}, 2021.

\bibitem{zhao2021fractional}
X.~Zhao, R.~Tao, W.~Li, W.~Philips, and W.~Liao, ``Fractional gabor
  convolutional network for multisource remote sensing data classification,''
  \emph{IEEE Transactions on Geoscience and Remote Sensing}, 2021.

\bibitem{hang2020classification}
R.~Hang, Z.~Li, P.~Ghamisi, D.~Hong, G.~Xia, and Q.~Liu, ``Classification of
  hyperspectral and lidar data using coupled cnns,'' \emph{IEEE Transactions on
  Geoscience and Remote Sensing}, vol.~58, no.~7, pp. 4939--4950, 2020.

\bibitem{hazirbas2016fusenet}
C.~Hazirbas, L.~Ma, C.~Domokos, and D.~Cremers, ``Fusenet: Incorporating depth
  into semantic segmentation via fusion-based cnn architecture,'' in
  \emph{Asian conference on computer vision}.\hskip 1em plus 0.5em minus
  0.4em\relax Springer, 2016, pp. 213--228.

\bibitem{audebert2018beyond}
N.~Audebert, B.~Le~Saux, and S.~Lef{\`e}vre, ``Beyond rgb: Very high resolution
  urban remote sensing with multimodal deep networks,'' \emph{ISPRS Journal of
  Photogrammetry and Remote Sensing}, vol. 140, pp. 20--32, 2018.

\bibitem{chen2019effective}
K.~Chen, K.~Fu, X.~Gao, M.~Yan, W.~Zhang, Y.~Zhang, and X.~Sun, ``Effective
  fusion of multi-modal data with group convolutions for semantic segmentation
  of aerial imagery,'' in \emph{IGARSS 2019-2019 IEEE International Geoscience
  and Remote Sensing Symposium}.\hskip 1em plus 0.5em minus 0.4em\relax IEEE,
  2019, pp. 3911--3914.

\bibitem{khan2020survey}
A.~Khan, A.~Sohail, U.~Zahoora, and A.~S. Qureshi, ``A survey of the recent
  architectures of deep convolutional neural networks,'' \emph{Artificial
  Intelligence Review}, vol.~53, no.~8, pp. 5455--5516, 2020.

\bibitem{zhang2019differentiable}
Z.~Zhang, J.~Li, W.~Shao, Z.~Peng, R.~Zhang, X.~Wang, and P.~Luo,
  ``Differentiable learning-to-group channels via groupable convolutional
  neural networks,'' in \emph{Proceedings of the IEEE/CVF International
  Conference on Computer Vision}, 2019, pp. 3542--3551.

\bibitem{krizhevsky2012imagenet}
A.~Krizhevsky, I.~Sutskever, and G.~E. Hinton, ``Imagenet classification with
  deep convolutional neural networks,'' \emph{Advances in neural information
  processing systems}, vol.~25, pp. 1097--1105, 2012.

\bibitem{xie2017aggregated}
S.~Xie, R.~Girshick, P.~Doll{\'a}r, Z.~Tu, and K.~He, ``Aggregated residual
  transformations for deep neural networks,'' in \emph{Proceedings of the IEEE
  conference on computer vision and pattern recognition}, 2017, pp. 1492--1500.

\bibitem{yin2019understanding}
P.~Yin, J.~Lyu, S.~Zhang, S.~Osher, Y.~Qi, and J.~Xin, ``Understanding
  straight-through estimator in training activation quantized neural nets,'' in
  \emph{International Conference on Learning Representations}, 2019.

\bibitem{he2016deep}
K.~He, X.~Zhang, S.~Ren, and J.~Sun, ``Deep residual learning for image
  recognition,'' in \emph{Proceedings of the IEEE conference on computer vision
  and pattern recognition}, 2016, pp. 770--778.

\bibitem{ronneberger2015u}
O.~Ronneberger, P.~Fischer, and T.~Brox, ``U-net: Convolutional networks for
  biomedical image segmentation,'' in \emph{International Conference on Medical
  image computing and computer-assisted intervention}.\hskip 1em plus 0.5em
  minus 0.4em\relax Springer, 2015, pp. 234--241.

\bibitem{okujeni2016berlin}
A.~Okujeni, S.~van~der Linden, and P.~Hostert, ``Berlin-urban-gradient dataset
  2009-an enmap preparatory flight campaign,'' 2016.

\bibitem{hong2021multimodal}
D.~Hong, J.~Hu, J.~Yao, J.~Chanussot, and X.~X. Zhu, ``Multimodal remote
  sensing benchmark datasets for land cover classification with a shared and
  specific feature learning model,'' \emph{ISPRS Journal of Photogrammetry and
  Remote Sensing}, vol. 178, pp. 68--80, 2021.

\bibitem{du2017technical}
\BIBentryALTinterwordspacing
X.~Du and A.~Zare, ``Technical report: scene label ground truth map for muufl
  gulfport data set,'' University of Florida, Gainesville, FL., Tech. Rep.
  Tech. Rep. 20170417, April 2017. [Online]. Available:
  \url{http://ufdc.ufl.edu/IR00009711/00001}
\BIBentrySTDinterwordspacing

\bibitem{cerra2018combining}
D.~Cerra, M.~Pato, E.~Carmona, S.~M. Azimi, J.~Tian, R.~Bahmanyar, F.~Kurz,
  E.~Vig, K.~Bittner, C.~Henry \emph{et~al.}, ``Combining deep and shallow
  neural networks with ad hoc detectors for the classification of complex
  multi-modal urban scenes,'' in \emph{IGARSS 2018-2018 IEEE International
  Geoscience and Remote Sensing Symposium}.\hskip 1em plus 0.5em minus
  0.4em\relax IEEE, 2018, pp. 3856--3859.

\bibitem{kingma2014adam}
D.~P. Kingma and J.~Ba, ``Adam: A method for stochastic optimization,''
  \emph{arXiv preprint arXiv:1412.6980}, 2014.

\bibitem{he2015delving}
K.~He, X.~Zhang, S.~Ren, and J.~Sun, ``Delving deep into rectifiers: Surpassing
  human-level performance on imagenet classification,'' in \emph{Proceedings of
  the IEEE international conference on computer vision}, 2015, pp. 1026--1034.

\bibitem{goyal2017accurate}
P.~Goyal, P.~Doll{\'a}r, R.~Girshick, P.~Noordhuis, L.~Wesolowski, A.~Kyrola,
  A.~Tulloch, Y.~Jia, and K.~He, ``Accurate, large minibatch sgd: Training
  imagenet in 1 hour,'' \emph{arXiv preprint arXiv:1706.02677}, 2017.

\bibitem{lecun1990optimal}
Y.~LeCun, J.~S. Denker, and S.~A. Solla, ``Optimal brain damage,'' in
  \emph{Advances in neural information processing systems}, 1990, pp. 598--605.

\bibitem{goodfellow2016deep}
I.~Goodfellow, Y.~Bengio, and A.~Courville, \emph{Deep learning}.\hskip 1em
  plus 0.5em minus 0.4em\relax MIT press, 2016.

\bibitem{kukavcka2017regularization}
J.~Kuka{\v{c}}ka, V.~Golkov, and D.~Cremers, ``Regularization for deep
  learning: A taxonomy,'' \emph{arXiv preprint arXiv:1710.10686}, 2017.

\bibitem{zheng2021deep}
Z.~Zheng, A.~Ma, L.~Zhang, and Y.~Zhong, ``Deep multisensor learning for
  missing-modality all-weather mapping,'' \emph{ISPRS Journal of Photogrammetry
  and Remote Sensing}, vol. 174, pp. 254--264, 2021.

\end{thebibliography}

% \end{thebibliography}


% \end{IEEEbiography}


% % insert where needed to balance the two columns on the last page with
% %\newpage



%\vfill


\end{document}